\documentclass{article}

\usepackage[numbers,sort&compress]{natbib}
\setcitestyle{numbers,sort&compress,square,comma}

\usepackage{amssymb}

\usepackage[preprint]{neurips_2025}

\usepackage{authblk} 

\usepackage[utf8]{inputenc} 
\usepackage[T1]{fontenc}    
\usepackage{newunicodechar}
\newunicodechar{∘}{\circ}
\usepackage[colorlinks=true, citecolor=green]{hyperref}       
\usepackage{url}            
\usepackage{booktabs}       
\usepackage{amsfonts}       
\usepackage{nicefrac}       
\usepackage{microtype}      
\usepackage{pifont}
\usepackage{array}
\usepackage{graphicx} 
\usepackage{siunitx}
\usepackage{multirow}
\usepackage{type1cm}
\usepackage{tabularx}        
\usepackage{enumitem}
\usepackage{amsmath}
\usepackage{floatrow}
\usepackage[table]{xcolor} 
\usepackage{adjustbox}
\usepackage{soul} 

\title{TDVE-Assessor: Benchmarking and Evaluating the Quality of Text-Driven Video Editing with LMMs}

\author{%
  Juntong Wang\textsuperscript{1}, Jiarui Wang\textsuperscript{1}, Huiyu Duan\textsuperscript{1}, Guangtao Zhai\textsuperscript{1}, Xiongkuo Min\textsuperscript{1} \\
  \textsuperscript{1}Institute of Image Communication and Network Engineering\\ 
  Shanghai Jiao Tong University, Shanghai, China\\
}

\begin{document}

\maketitle

\begin{abstract}
Text-driven video editing is rapidly advancing, yet its rigorous evaluation remains challenging due to the absence of dedicated video quality assessment (VQA) models capable of discerning the nuances of editing quality. To address this critical gap, we introduce \textbf{TDVE-DB}, a large-scale benchmark dataset for \underline{t}ext-\underline{d}riven \underline{v}ideo \underline{e}diting. TDVE-DB consists of 3,857 edited videos generated from 12 diverse models across 8 editing categories, and is annotated with 173,565 human subjective ratings along three crucial dimensions, \textit{i.e.}, edited video quality, editing alignment, and structural consistency. Based on TDVE-DB, we first conduct a comprehensive evaluation for the  12 state-of-the-art editing models revealing the strengths and weaknesses of current video techniques, and then benchmark existing VQA methods in the context of text-driven video editing evaluation. Building on these insights, we propose \textbf{TDVE-Assessor}, a novel VQA model specifically designed for text-driven video editing assessment. TDVE-Assessor integrates both spatial and temporal video features into a large language model (LLM) for rich contextual understanding to provide comprehensive quality assessment. Extensive experiments demonstrate that TDVE-Assessor substantially outperforms existing VQA models on TDVE-DB across all three evaluation dimensions, setting a new state-of-the-art. Both TDVE-DB and TDVE-Assessor will be released upon the publication.
\end{abstract}

\section{Introduction}

With the explosive growth of artificial intelligence generated content (AIGC), text-driven video editing technologies \cite{tune-a-video,tokenflow,text2video-zero,ccedit,controlvideo,fatezero,flatten,fresco,pix2video,rave,slicedit,vid2vid} have emerged as a critical bridge connecting human creativity with machine intelligence. Users can now achieve complex modifications to video content through simple textual instructions. However, this powerful capability also introduces unprecedented evaluation challenges: How can we objectively and comprehensively measure whether these models truly understand and accurately execute the users' editing intent? Existing models \cite{tune-a-video,tokenflow,vid2vid,text2video-zero} often produce semantic deviations and spatio-temporal artifacts when facing complex scenarios (\textit{e.g.}, multi-object interactions, fine-grained attribute adjustments), while traditional video quality assessment (VQA) methods \cite{zongshu,dover,fastvqa,simplevqa,VSFA,BVQA} and existing benchmark models \cite{vebench,T2VQA,AIGVASSESSOR} fall short or align poorly with human preferences in handling such multimodal, fine-grained editing tasks. This gap in evaluation capabilities has become a critical bottleneck constraining the further progress of the field.

To break this end, we first construct \textbf{TDVE-DB}, currently the largest and most comprehensive dataset \cite{ccedit,vebench,TGVE} for \underline{t}ext-\underline{d}riven \underline{v}ideo \underline{e}diting quality assessment, featuring the broadest range of editing categories. TDVE-DB consists of 3,857 edited videos produced by 12 state-of-the-art open-source text-driven video editing models using 340 editing prompts covering from 8 different editing perspectives (\textit{e.g.}, object replacement, style transfer). Based on these source videos, editing prompts, and edited videos, we conduct rigorously controlled subjective assessment experiments and collect 173,565 subjective ratings. These ratings cover three critical dimensions for video editing: (1) \textbf{edited video quality}, (2) \textbf{editing alignment} (\textit{i.e.}, the alignment between the edited result and the prompt instruction), and (3) \textbf{structural consistency} (\textit{i.e.}, the preservation of structure and content from the source video in the edited video).

Furthermore, we propose \textbf{TDVE-Assessor}, a large multimodal model (LMM) \cite{llmzongshu}  based video quality \underline{assess}ment method, specifically designed for text-driven video editing (\underline{TDVE}).TDVE-Assessor introduces an innovative approach by reframing the quality assessment task as an interactive question-answering (QA) framework. It leverages the powerful multimodal understanding and reasoning capabilities of LMMs to not only generate interpretable qualitative evaluations but also to output precise quantitative scores through a fine-tuned score regression module. Extensive experimental results demonstrate that TDVE-Assessor significantly outperforms various existing evaluation methods on our constructed TDVE-DB and other public benchmarks \cite{vebench,T2VQA}. Our main contributions are summarized as follows.

\begin{itemize}[itemsep=0pt,parsep=0pt,topsep=0pt,partopsep=0pt,
               leftmargin=*,    
               labelindent=0pt] 
\item  We construct \textbf{TDVE-DB, currently the largest text-driven video editing quality assessment benchmark dataset}, featuring the most diverse range of editing categories and fine-grained multi-dimensional human subjective ratings. This dataset provides a solid foundation for comprehensively evaluating existing video editing models.
\item  We propose \textbf{TDVE-Assessor}, a novel LMM-based video quality assessment model. TDVE-Assessor integrates both spatial and temporal video features into a large language model (LLM) for rich contextual understanding to provide comprehensive quality assessment (edited video quality, editing alignment, and structural consistency) for text-driven video edits.
\item  Our experiments demonstrate that TDVE-Assessor achieves \textbf{state-of-the-art (SOTA) performance} on the TDVE-DB dataset and shows excellent generalization capabilities across several other benchmarks, validating its effectiveness and robustness for promoting future research in video editing quality assessment.
\end{itemize}

\begin{figure}[t]
\vspace{-0.8cm}
\setlength{\belowcaptionskip}{-0.7cm}
\centering   
\includegraphics[width=\textwidth]{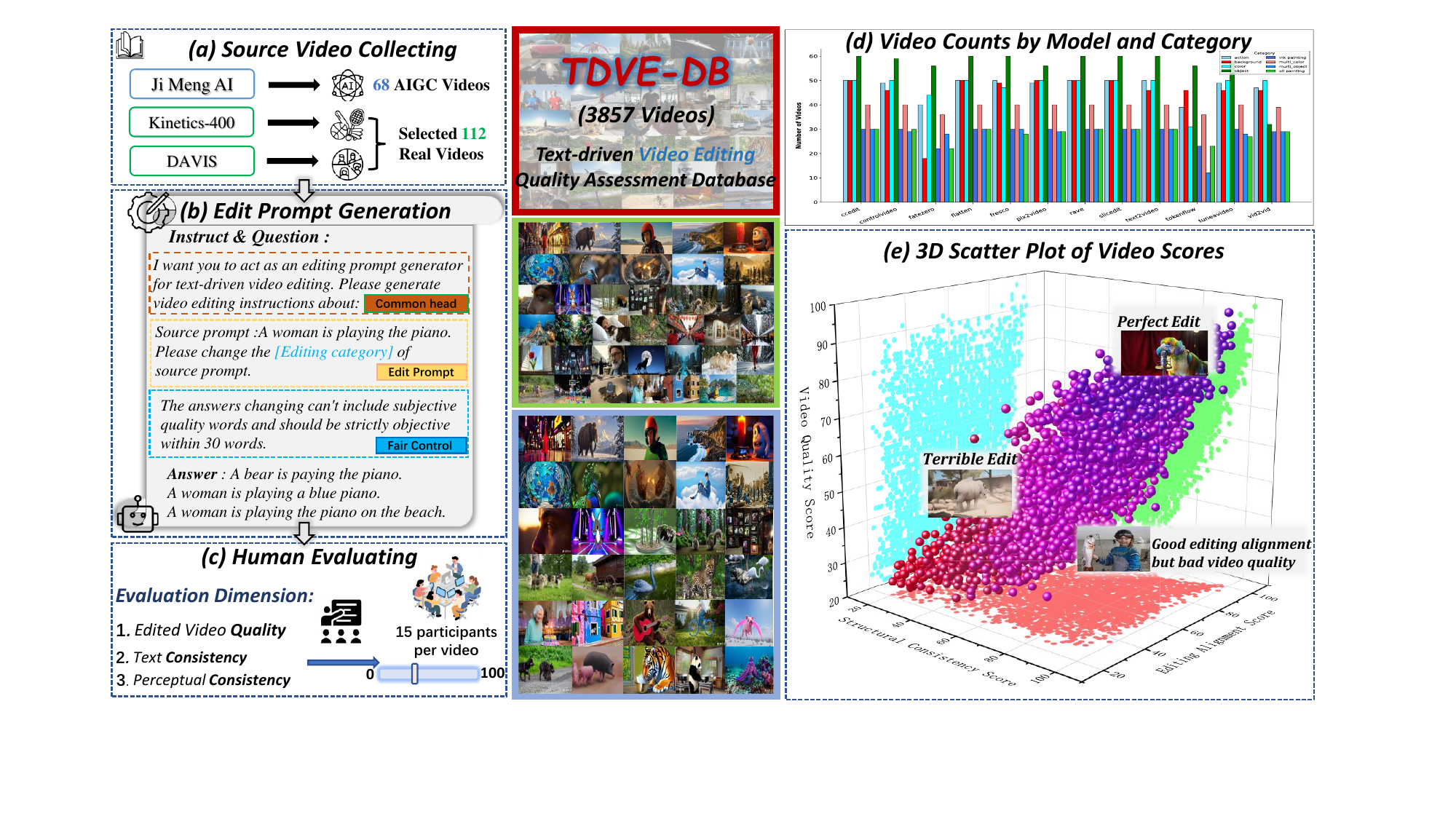} 

\caption{(a) Acquisition of the source video
(b) Generation of prompt words
(c) Obtaining 170K subjective scores through subjective experiments
(d) The number of videos for different models and different editing categories. 
(e) Three-dimensional scatter plot of subjective scores} 
\label{dataset}
\end{figure}
\vspace{-0.1cm}

\section{Related Work}
\label{gen_inst}
\subsection{Text-Driven Video Editing Model}
\label{subsec:Text-Driven Video Editing Model}
The field of text-driven video editing has witnessed rapid advancements, largely driven by the emergence of diffusion-based generative models. Early methods such as Tune-A-Video~\cite{tune-a-video}, introduce single-sample fine-tuning with spatiotemporal attention mechanisms, while TokenFlow~\cite{tokenflow} enhances temporal consistency through diffusion feature propagation. Text2Video-Zero~\cite{text2video-zero} further advances the field by enabling zero-shot long-form video generation. CCEdit \cite{ccedit} achieves precise editing through structure-appearance disentanglement, ControlVideo~\cite{controlvideo} utilizes full cross-frame attention for coherent temporal manipulation. FateZero~\cite{fatezero} preserves motion dynamics via reverse attention fusion. Pix2Video~\cite{pix2video} mitigates flickering artifacts through self-attention feature injection. Recent advancements prioritize zero-shot optimization. RAVE~\cite{rave} enhances memory efficiency for longer sequences, Vid2Vid-Zero~\cite{vid2vid} tackles motion inconsistencies, and FLATTEN~\cite{flatten} introduced optical flow-guided attention for maintaining dynamic consistency. FRESCO~\cite{fresco} imposes dual spatiotemporal constraints to stabilize output, while SlicEdit~\cite{slicedit} achieves high visual fidelity through slice-based preprocessing strategies. Despite these advancements, evaluating the diverse outputs of these models poses a significant challenge.
\begin{table}[t]
  \label{video editing model}
  \caption{Overview of text-driven video editing models used for dataset construction, detailing key parameters.The phrase ‘Follow Source’ means that the output video maintains the same resolution as the source video. SD means Stable Diffusion. }
  \label{tab:example1}
  \centering
  \resizebox{\textwidth}{!}{
  \begin{tabular}{cccccccc}  
    \toprule
    \textbf{Model} & \textbf{Year} & \textbf{Length} & \textbf{Base Model} & Resolution & \textbf{FPS} &  \textbf{Zero-shot} & \textbf{Open Source} \\
    \midrule
    Tune-A-Video \cite{tune-a-video}  & 22.12 & 3s & 512$\times$512 & SD 1-4 & 8  &\ding{55}   & \checkmark \\
    Tokenflow \cite{tokenflow}    & 23.07 & 1s & 512$\times$512 & SD 2-1  & 30  & \checkmark   & \checkmark \\
    Text2Video-Zero \cite{text2video-zero} &23.03 & 1-5s & Follow Source & SD 1-5 & 24  &\checkmark  & \checkmark \\
    CCEdit \cite{ccedit}       & 23.09 & 2s & 768$\times$512 & SD 1-5 & 6  &\checkmark  & \checkmark \\
    ControlVideo \cite{controlvideo} & 23.05 & 1s & 512$\times$512 & SD 1-5 & 8  &\checkmark  & \checkmark \\
    FateZero \cite{fatezero}     & 23.03 & 1-4s & 512$\times$512 & SD 1-4 & 10  &\checkmark  & \checkmark \\
    FLATTEN \cite{flatten}      & 23.12 & 1-2s & 512$\times$512 & SD 2-1 & 15  &\checkmark  & \checkmark \\
    FRESCO  \cite{fresco}      & 24.06 & 1-5s & Follow Source & SD 1-5 & 24  &\checkmark  & \checkmark \\
    Pix2Video  \cite{pix2video}   & 23.03 & 2-4s & 512$\times$512 & SD 2 & 30  &\checkmark  & \checkmark \\
    RAVE \cite{rave}         & 23.12 & 1-3s & Follow Source & SD 1-5 & 30  &\checkmark  & \checkmark \\
    Slicedit  \cite{slicedit}    & 24.05 & 1-4s & Follow Source & SD 1-5 & 25  &\checkmark  & \checkmark \\
    vid2vid-zero \cite{vid2vid} & 23.03 & 2s & 512$\times$512 & SD 2-1 & 8  &\checkmark  & \checkmark \\
    \bottomrule
  \end{tabular}
  }
\end{table}

\subsection{Datasets for Video Editing Quality Assessment}
\label{subsec:Datasets for Video Editing Quality Assessment}
While numerous video datasets have been proposed, resources specifically designed for video editing quality assessment remain limited. Señorita-2M \cite{senorita2m} offers 2 million edited pairs but lacks Mean Opinion Score (MOS) annotations for video quality. Smaller datasets such as TGVE \cite{TGVE} and BalanceCC \cite{ccedit} have limited scale and categorical granularity. VE-Bench DB \cite{vebench} provides MOS scores, but its utility is constrained by the single type of edit, yielding only 1,170 pairs. These limitations highlight the need for a more comprehensive benchmark.
To address these gaps, we introduce \textbf{TDVE-DB}, a new benchmark dataset comprising  180 diverse source videos (real-world \cite{davis,kinetics} and AI-generated \cite{jimengAI}), 3,857 edited pairs from 12 state-of-the-art models covering 8 editing categories (\textit{e.g.}, object replacement, style transfer), and MOSs from human evaluators following ITU-T P.910 \cite{p910}, making it a robust resource for the field.

\begin{table}[t]
  \caption{Comparison of Existing Video Editing Datasets with the Proposed TDVE-DB, Focusing on Key Metrics and Features.}
  \label{tab:example}
  \centering
  \resizebox{\textwidth}{!}{
  \begin{tabular}{ccccccccc}  
    \toprule
    \textbf{Dataset} & \textbf{Samples} & \textbf{Edited} & \textbf{MOS} & \textbf{Edit Dimension}& \textbf{Evaluate Dimension} & \textbf{Count of video editing models} & \textbf{FPS} & \textbf{Open Source} \\
    \midrule
       Señorita-2M \cite{senorita2m} & 2M & \checkmark & \ding{55} &\ding{55} & \ding{55} & 4 & $\geq 8$  & \checkmark \\
       LOVEU-TGVE-2023 \cite{TGVE} & 78 & \ding{55} & \ding{55} & \ding{55} & \ding{55} & \ding{55} & 8 & \checkmark \\
       BalanceCC \cite{ccedit} & 412 & \checkmark & \ding{55} & 4 & \ding{55} & 1 & 8 & \checkmark \\
       VE-Bench DB \cite{vebench} & 1170 & \checkmark & $\approx 28,000$ & 3 & 1 & 8 & 8 & \checkmark \\
       \midrule
       \textbf{TDVE-DB} & 3857 & \checkmark & 173,565 & 8 & 3 & 12 & 6-30 & \checkmark \\
    \bottomrule
  \end{tabular}
  }
    \label{VE-dataset}
\end{table}

\subsection{Video Quality Assessment Model}
\label{subsec:Video Quality Assessment Model}
The field of Video Quality Assessment (VQA) encompasses a diverse array of models, ranging from traditional handcrafted feature-based methods (\textit{e.g.}, HOST, QAC, NIQE \cite{HOST,qac,niqe}) to deep learning-based VQA approaches (\textit{e.g.}, DOVER \cite{dover}, FAST-VQA \cite{fastvqa}, BVQA \cite{BVQA}). While these models excel at predicting structural quality scores by characterizing quality-aware information, they often struggle to evaluate the crucial text-to-video (T2V) correspondence, which is vital for assessing the relationship between generated/edited videos and their corresponding textual prompts. Although text-video correspondence metrics like CLIPScore \cite{CLIPScore} and VQAScore \cite{vqascore} have improved the evaluation in this regard, they still face challenges in providing effective multi-dimensional assessments for video editing tasks (\textit{e.g.}, evaluating visual quality of edits, adherence to text instructions, and consistency with original content simultaneously). Furthermore, while Large Multimodal Models (LMMs) \cite{llmzongshu,lmm4lmm} demonstrate strong semantic understanding, they typically lack the capability to output precise, quantifiable quality scores. Existing efforts tailored to video editing evaluation, such as VE-Bench QA \cite{vebench}, though pioneering, have limitations in terms of the comprehensiveness of evaluation dimensions and generalization capabilities across diverse editing types and models. To address these shortcomings, we propose \textbf{TDVE-Assessor}, a systematic and integrated video editing evaluation model designed to deliver accurate and effective assessments across multiple key facets of edited videos.

\section{Dataset Acquisition}
\label{sub:Dataset Acquisition}
\subsection{Data Collection}
\label{subsec:Data Collection}
\textbf{Video Sources.} For broad generalizability, our video sources include AI-generated content and real-world footage. We used ByteDance's Jimeng AI \cite{jimengAI} to create 68 high-definition (5s) synthetic videos tailored for editing, covering diverse semantic categories and dynamic interactions. Additionally, 112 real-world clips are curated from Kinetics-400 \cite{kinetics} and DAVIS \cite{davis} (50\% human actions, 15\% animal behaviors, 35\% others). This yielded 180 base videos with balanced domain representation.

\textbf{Editing Prompts.} We target mainstream editing tasks with a prompt system spanning eight aspects: color, motion, background, object manipulation, multi-color fusion, multi-object interaction, and artistic stylizations (oil painting, ink-ish). Using DeepSeek-R1 \cite{deepseekR1}, we generated 340 instructions via constrained prompt engineering. Each prompt aimed to modify specific attributes while preserving 60\% original content semantics, ensuring focused edits without unintended alterations.

\textbf{Video Editing Models.} For comprehensive evaluation, we select 12 open-source, diffusion-based video editing models: Tune-A-Video \cite{tune-a-video}, TokenFlow \cite{tokenflow}, Text2Video-Zero \cite{text2video-zero}, CCEdit \cite{ccedit}, ControlVideo \cite{controlvideo}, FateZero \cite{fatezero}, FLATTEN \cite{flatten}, FRESCO~\cite{fresco}, Pix2Video \cite{pix2video}, RAVE \cite{rave}, SlicEdit \cite{slicedit}, and Vid2Vid-Zero \cite{vid2vid}. Most (10/12) operate in zero-shot mode, all leveraging Stable Diffusion. Dual validation yielded 3,857 valid edited video pairs.

\subsection{Video Editing Evaluation Dimensions}
\label{subsec:Video Editing Evaluation Dimensions}
Our dataset, designed for text-driven video editing models, employs three core evaluation dimensions.
\textbf{Edited video quality} assesses the standalone visual quality of edited videos, akin to no-reference VQA \cite{zongshu}.
\textbf{Editing alignment} assesses how faithfully the edited video executes the semantic instructions in the edit prompt. In our subjective experiments, to ensure objective and quantifiable ratings, evaluators are guided to primarily focus on whether the explicit semantic changes indicated by the edit prompt are accurately implemented relative to the source video (often contextualized by its description). This emphasis on “delta changes” help establish a consistent judgment standard. While TDVE-Assessor primarily uses the edit prompt as its textual input for this dimension, it is trained on a large corpus of (edit prompt, edited video, human subjective scores reflecting this “change-centric” evaluation) tuples. Through this data-driven process, the model implicitly learns to discern the successful execution of an edit, mirroring human sensitivity to whether key instructed modifications are present and accurate in the output video.
\textbf{Structural consistency} measures spatial-structural similarity and visual continuity between source and edited videos.
These dimensions provide a comprehensive score set for each video pair, enabling systematic benchmarking of model fidelity, prompt adherence, and content preservation.

\begin{figure}[t]
\centering 
\setlength{\abovecaptionskip}{-0.2cm} 
\setlength{\belowcaptionskip}{-0.5cm} 
\includegraphics[width=\textwidth]{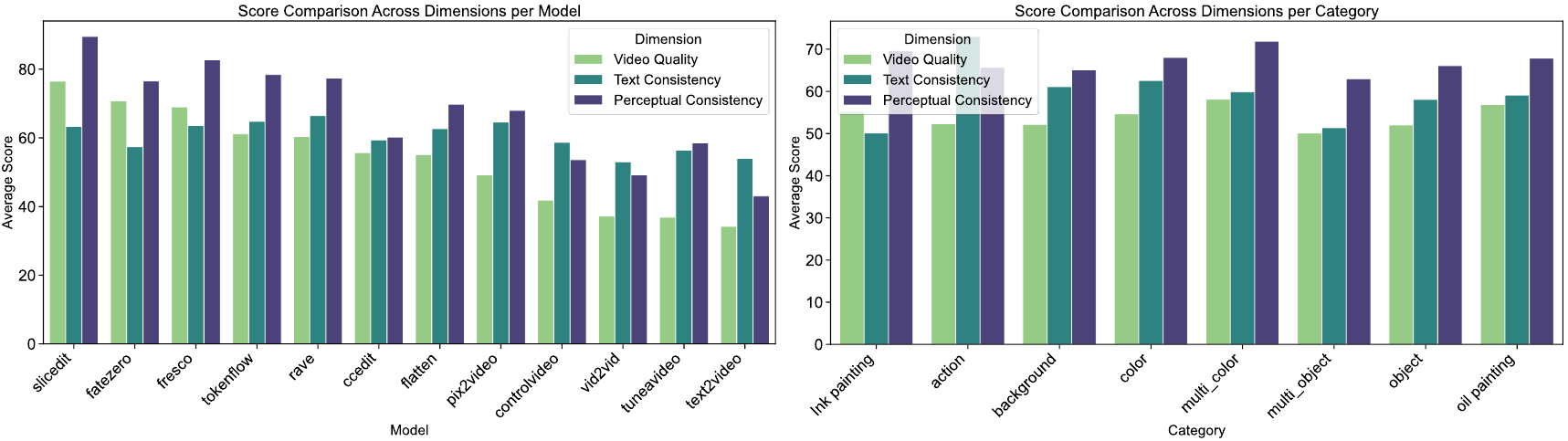} 
\caption{Bar charts of the score distribution of each model and editing category across three dimensions.}  
\label{score_zhu}
\end{figure}

\subsection{Subjective Video Editing Quality Assessment}
\label{subsec:Subjective Video Editing Quality Assessment}
We recruit 15 volunteers (good vision, balanced gender, aged 22-35 years, all of whom are either currently enrolled postgraduate students or possess postgraduate degrees) who view videos on a 27" pro monitor (2560×1440@120Hz, 80cm viewing distance.) meeting international standards. The three-phase experiment include: (1) Calibration Check: Participants pass a 35-video quiz, matching expert ratings by $\geq 85\%$. (2) Skill Building: Training involves 10 real-case comparisons per scoring category (\textit{e.g.}, color, motion). (3) Real Scoring: 480 video pairs are randomly presented, with 24 hidden repeats for reliability checks.

Consistent criteria are applied for video quality and structural consistency across categories. For textual consistency, the primary variable we use category-specific emphases and tiered grading. For example, multi-object scenarios had three tiers for object replacement accuracy (complete, partial, failure). Oil-painting style videos are first tiered by stylistic adherence, then scored refinedly within tiers considering other descriptors. The experiment took 2 hours per participant per time, with USD 13.5 compensation. Every participant should participate 8 times.

\subsection{Dataset Statistics and Analysis}
\label{subsec:Dataset Statistics and Analysis}

\begin{figure}[t]
\vspace{-0.2cm}
\centering 
\setlength{\abovecaptionskip}{-0.5cm} 
\setlength{\belowcaptionskip}{-0.5cm}

\includegraphics[width=\textwidth]{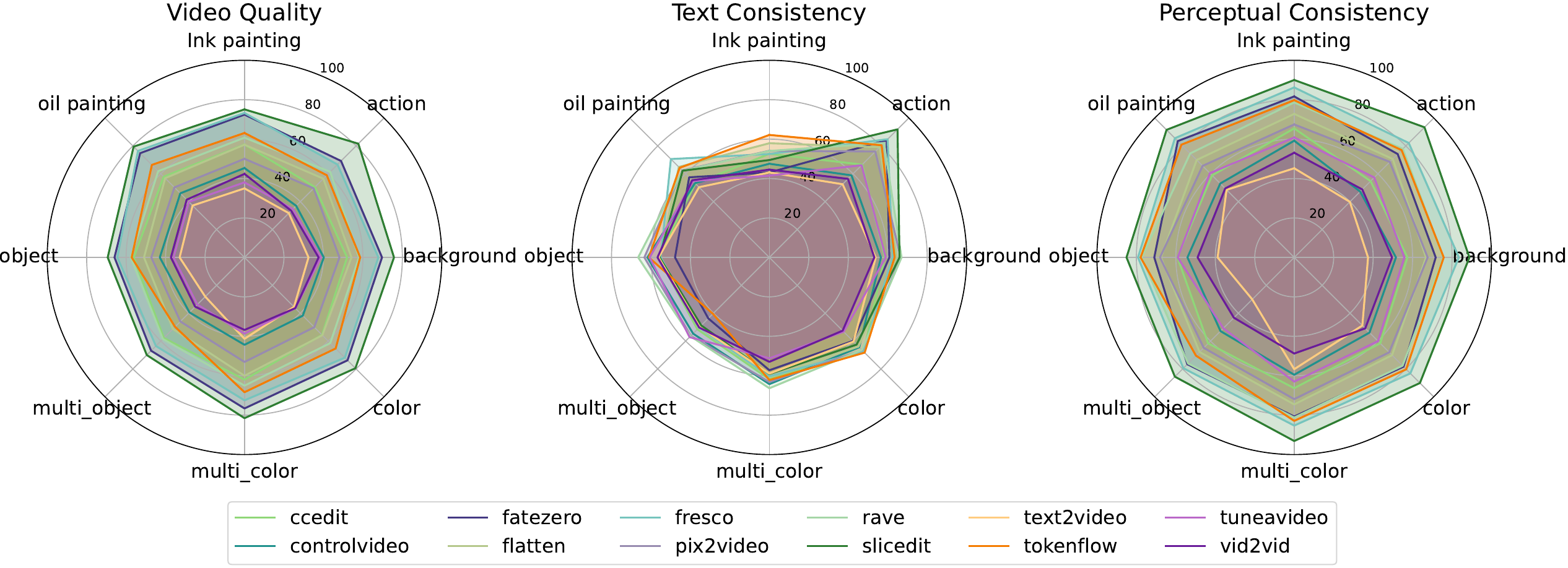} 
\caption{Radar charts of the score distribution of each model and editing category across three dimensions.} 
\label{score_leida}
\end{figure}

\begin{figure}[t]
\vspace{-0.7cm}
\centering 
\includegraphics[width=\textwidth]{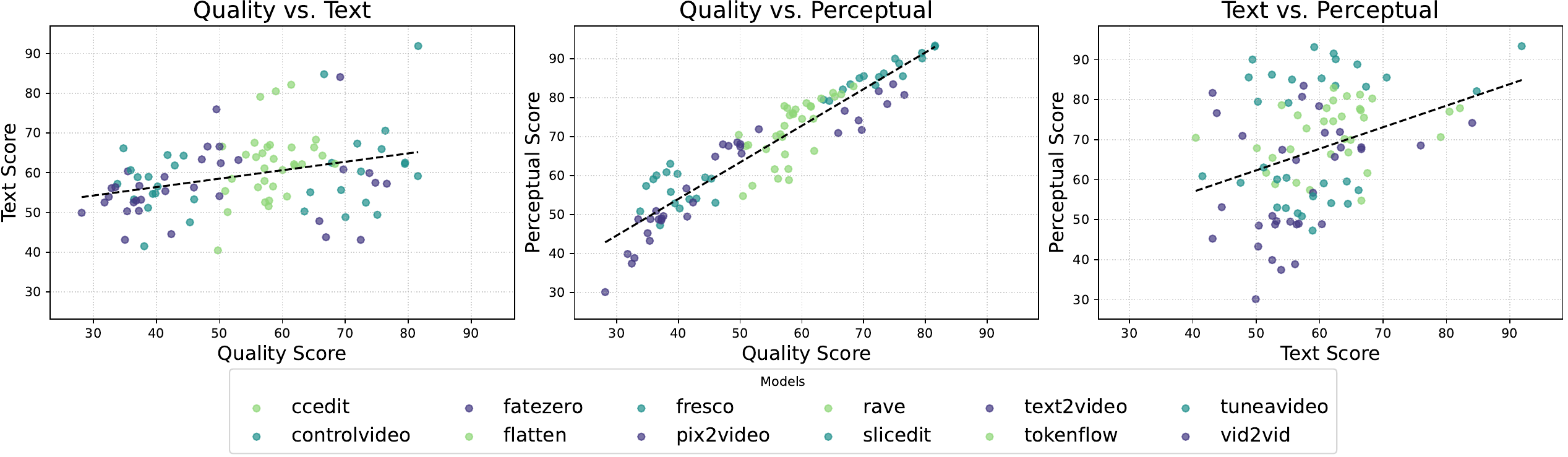} 
\caption{Scatter plots of score distribution between pairwise dimensions.} 
\label{weidu_sandian}
\end{figure}

After obtaining the subjective scores from the experimenters, we first conducted a consistency analysis of these subjective scores, and the results are quite good. The specific analysis can be found in the appendix \ref{subsec:icc_analysis}. To ensure the reliability and comparability of subjective scores, we processed the raw human ratings to derive MOS. The process is as follows:
\begin{subequations}
\label{eq:mos_calculation_group_v2}
\begin{equation}
    \label{eq:z_scores_and_params_v2}
    z_{ij} = \frac{r_{ij} - \mu_i}{\sigma_i}, \quad z'_{ij} = \frac{100(z_{ij} + 3)}{6}, \quad \mu_i = \frac{1}{N_i} \sum_{j=1}^{N_i} r_{ij}, \quad \sigma_i = \sqrt{\frac{1}{N_i-1} \sum_{j=1}^{N_i} (r_{ij} - \mu_i)^2}
\end{equation}
\begin{equation}
    \label{eq:mos_final_definition_v2}
    MOS_j = \frac{1}{M} \sum_{i=1}^{M} z'_{ij}
\end{equation}
\end{subequations}
where $r_{ij}$ is the raw rating from subject $i$ for video $j$. $N_i$ is the number of videos judged by subject $i$. $\mu_i$ and $\sigma_i$ are the mean and standard deviation of subject $i$'s ratings, respectively. $z_{ij}$ is the Z-score, $z'_{ij}$ is the scaled Z-score, and $MOS_j$ is the final Mean Opinion Score for video $j$ averaged over $M$ subjects. This process yield a robust MOS for each video across the evaluated dimensions, forming the basis for our subsequent analysis.
Figure \ref{score_zhu} analyzes 12 models across 8 editing categories on three dimensions. For video quality, Slicedit \cite{slicedit}, FRESCO \cite{fresco}, and FateZero \cite{fatezero} excelled, especially in color and style alterations.
In editing alignment, RAVE \cite{rave} led, followed by Tokenflow \cite{tokenflow} and Pix2video \cite{pix2video}. Among the various editing categories, color editing is a top-performing category. Single-entity edits (\textit{e.g.}, “color”) generally scored higher than multi-entity edits (\textit{e.g.}, “multi-color”).
For structural consistency (correlation with source structure), Slicedit \cite{slicedit} and FRESCO \cite{fresco} again excelled, with color edits also performing well in this dimension. This trend, mirroring video quality, likely stems from high-quality source videos where preserving structure yields comparable quality, thus correlating these dimensions. This also explains the strong performance of traditional VQA models \cite{dover,fastvqa,BVQA} on this dimension in later benchmarks.
Radar charts (Figure \ref{score_leida}) offer clearer dimensional comparisons, and pairwise scatter plots (Figure \ref{weidu_sandian}) confirm the positive correlation between video quality and structural consistency. These findings provide valuable insights into current text-driven video editing capabilities.

\section{Proposed Method}
\label{sec:proposed_method}

\begin{figure}[t]
\centering
\includegraphics[width=\textwidth]{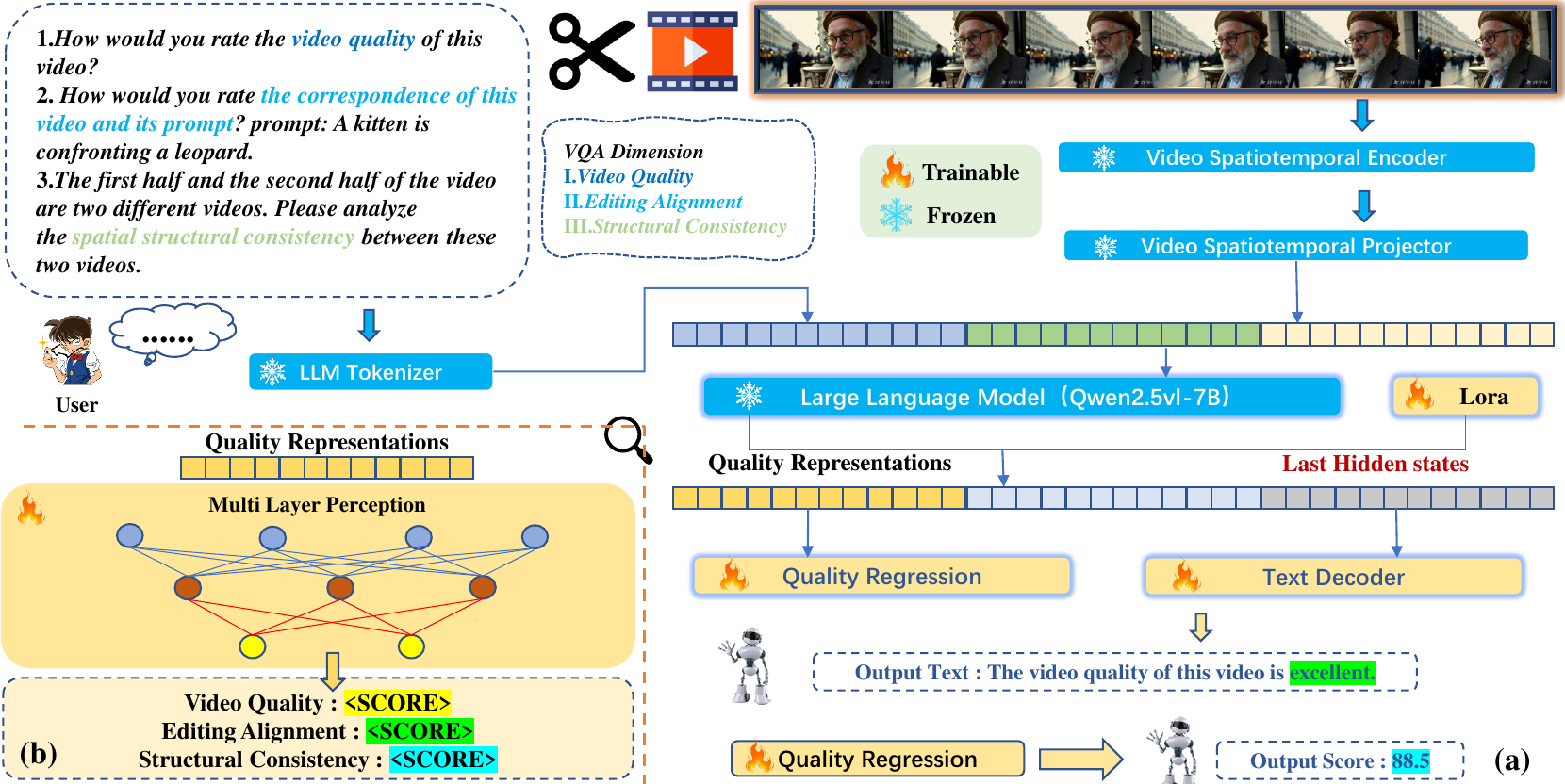} 
\caption{ (a) Video features are extracted by a frozen spatiotemporal encoder and aligned to a Large Language Model (Qwen2.5-VL-7B) using trainable projection modules. The LLM, fine-tuned with LoRA, then generates textual feedback for multiple evaluation dimensions via a trainable text decoder, while its last hidden states (quality representations) are input to a quality regression module. The model also supports pairwise comparison fine-tuning. (b) The quality regression module, a trainable MLP, converts these LLM-derived quality representations into numerical scores for dimensions like video quality, editing alignment, and structural consistency.}
\label{model_struct}
\end{figure}

Our \textbf{TDVE-Assessor}, depicted in Figure \ref{model_struct}{}, comprises two main components for visual processing and subsequent quality evaluation.

\subsection{Model Structure}
\label{subsec:model_structure}

For \textbf{visual encoding}, we employ the pre-trained Visual Transformer (ViT) \cite{VIT1,VIT2,VIT3,VIT4,VIT5,VIT6,VIT7} module embedded in Qwen2.5-VL-7B-Instruct \cite{Qwen-VL} to extract visual features.A major advantage of this module lies in its inherent capability to process video frames with diverse resolutions and frame rates directly, eliminating the need for lossy preprocessing techniques such as resizing or compression, which can inadvertently discard crucial information for video quality assessment. The resulting visual features $F_v$ are then projected into a dedicated latent space $F'_v$ via a lightweight, native two-layer Multi-Layer Perceptron (MLP)\cite{mlp}, providing a unified representation space that facilitates effective multimodal fusion in downstream tasks.

The \textbf{quality assessment} is performed using Qwen2.5VL-7B \cite{Qwen-VL} as the LMM, which integrates the processed visual features $F'_v$ with relevant textual information. This framework is designed to produce a dual output for a comprehensive evaluation. Firstly, it performs \textit{Qualitative Description Generation}, where the LMM generates descriptive textual evaluations of video quality, such as “The facial quality in this video is \{bad|poor|fair|good|excellent\}.” This leverages the LMM's inherent text processing capabilities to provide intuitive, user-friendly feedback and can also serve as a preliminary classification step, potentially guiding the more granular regression task. Secondly, the framework handles \textit{Quantitative Score Regression}. In this subtask, a lightweight regression head operates on a pooled representation derived from the LMM's final hidden states (\textit{e.g.}, the hidden state corresponding to the first token), $H_{\text{LMM\_pool}} \in \mathbb{R}^{D_{\text{LMM}}}$, to predict precise, continuous quality scores $\hat{s}$. This regression head is an MLP structured as follows: a linear layer projects $H_{\text{LMM\_pool}}$ from the LMM's hidden dimension $D_{\text{LMM}}$ to an expanded intermediate dimension (specifically, $2 \times D_{\text{LMM}}$), followed by a GELU activation function, and a final linear layer maps this intermediate representation to a single scalar value representing the quality score. This dual-output design thus enables the model to evaluate videos both qualitatively for interpretability and quantitatively for precise, objective assessment.

\vspace{-0.3cm}
\subsection{Training and Fine-tuning Strategy}
\label{subsec:training_strategy}

\textbf{TDVE-Assessor} employs a dual-stage strategy: (1) \textbf{text-aligned instruction pretraining} for quality-level classification, followed by (2) \textbf{LoRA-enhanced fine-tuning} for continuous score regression.

\textbf{Text-Aligned Instruction Pretraining} bridges numerical MOS scores and LLM text capabilities. We discretize MOS scores into $N_L=5$ ITU-defined textual levels (\{\textit{bad|pool|fair|good|excellent}\}) using Eq.~\eqref{eq:discretization} to divide the $[m, M]$ range, facilitating text-based learning and mitigating class imbalance.
\begin{equation}
    L_{\text{map}}(s) = \text{label}_i, \quad \text{if } m + \frac{i-1}{N_L} (M-m) \le s < m + \frac{i}{N_L} (M-m)
    \label{eq:discretization}
\end{equation}
where $s$ is MOS score, $L_{\text{map}}(s)$ maps to the $i$-th label $\text{label}_i$, $m, M$ are min/max MOS. We integrate these levels with nine distortion categories (\textit{e.g.}, blur, noise) into an instruction dataset. Training uses a language modeling loss ($L_{\text{pretrain}}$) to jointly optimize quality-level prediction and distortion classification (when applicable), fostering quality-distortion correlations.

\textbf{LoRA-Enhanced Quality Regression Fine-tuning} \cite{LORA} aims to train the model to predict continuous MOS scores using the MLP-based regression head detailed in Section~\ref{subsec:model_structure}. We use LoRA \cite{LORA} for parameter-efficient fine-tuning of the LMM (updating only low-rank matrices $A, B$ where $\Delta W = BA$), while the lightweight regression head is trained fully. This stage minimizes the L1 loss (Mean Absolute Error) between the predicted scores $\hat{s}$ and the ground-truth MOS $s$. This optimizes for score regression accuracy while preserving the LMM's general capabilities. For more details, please refer to the appendix \ref{sec:appendix_training_details}.

\section{Experimental Validation}
\label{sec:experimental_validation}

We validate \textbf{TDVE-Assessor} on four datasets: our \textbf{TDVE-DB}, VE-Bench DB \cite{vebench} (both text-driven video editing quality benchmarks), the AIGV quality dataset T2VQA-DB \cite{T2VQA}, and the AIGV quality dataset AIGVQA-DB \cite{AIGVASSESSOR} . 

\begin{table}[t]
\centering
\caption{Performance Benchmark on TDVE-DB. $\clubsuit$ vision-language pre-training models, $\diamondsuit$ open-source LLM-based models , $\bigstar$ Conventional handcrafted metrics, 
$\spadesuit$ deep learning-based VQA models. The best results
are marked in \textcolor{red}{RED} and the second-best in \textcolor{blue}{BLUE}.
}

\footnotesize
\setlength{\tabcolsep}{2.5pt} 
\renewcommand{\arraystretch}{0.99} 

\resizebox{\textwidth}{!}{%
  \begin{tabular}{@{} 
    >{\hspace{0.5em}}p{4.5cm} 
    *{3}{S[table-format=1.4]} 
    *{3}{S[table-format=1.4]} 
    *{3}{S[table-format=1.4]} 
    *{3}{S[table-format=1.4]} 
  @{}}
  \toprule
  \multirow{2}{*}{Method} & 
  \multicolumn{3}{c}{\textbf{Video Quality}} & 
  \multicolumn{3}{c}{\textbf{Editing Alignment}} & 
  \multicolumn{3}{c}{\textbf{Structural Consistency}} &
  \multicolumn{3}{c}{\textbf{Overall Average}} \\
  \cmidrule(lr){2-4} \cmidrule(lr){5-7} \cmidrule(lr){8-10} \cmidrule(l){11-13}
  & {SRCC} & {PLCC} & {KRCC} & {SRCC} & {PLCC} & {KRCC} & {SRCC} & {PLCC} & {KRCC} & {SRCC} & {PLCC} & {KRCC} \\

  \midrule
  $\clubsuit$ ImageReward \cite{xu2023imagereward} & 0.0301 & 0.0179 & 0.0202 & 0.0312 & 0.0193 & 0.0210 & 0.0858 & 0.0736 & 0.0578 & 0.0490 & 0.0369 & 0.0330 \\
  $\clubsuit$ BLIPScore \cite{li2022blip} & 0.0911 & 0.0914 & 0.0610 & 0.0910 & 0.0915 & 0.0610 & 0.0408 & 0.0397 & 0.0271 & 0.0743 & 0.0742 & 0.0497 \\
  $\clubsuit$ CLIPScore \cite{CLIPScore}& 0.0173 & 0.0222 & 0.0114 & 0.1970 & 0.2086 & 0.1343 & 0.0264 & 0.0210 & 0.0185 & 0.0802 & 0.0839 & 0.0547 \\
  $\clubsuit$ PickScore \cite{Kirstain2023PickaPicAO}& 0.0335 & 0.0216 & 0.0228 & 0.3177 & 0.3084 & 0.2153 & 0.0037 & 0.0141 & 0.0026 & 0.1183 & 0.1147 & 0.0802 \\
  $\clubsuit$ VQAScore \cite{vqascore}& 0.0586 & 0.0339 & 0.0396 & 0.3761 & 0.3627 & 0.2613 & 0.1015 & 0.0830 & 0.0685 & 0.1787 & 0.1599 & 0.1231 \\
  $\clubsuit$ AestheticScore \cite{aesthe} & 0.2203 & 0.2112 & 0.1501 & 0.2224 & 0.2132 & 0.1515 & 0.2499 & 0.2341 & 0.1713 & 0.2309 & 0.2195 & 0.1576 \\
  \midrule
  $\diamondsuit$ LLava-NEXT  \cite{llavanext-video}& 0.1109 & 0.1111 & 0.0886 & 0.1446 & 0.1274  & 0.1185 &  0.0559  & 0.0540   & 0.0434 & 0.1038 & 0.0975 & 0.0835 \\
  $\diamondsuit$ InternVideo2.5 \cite{internvideo} & 0.1171 & 0.1364 & 0.0891 & 0.3650 & 0.3290  & 0.2922 & 0.6004 & 0.5806 & 0.4802 & 0.3608 & 0.3487 & 0.2872 \\
  $\diamondsuit$ VideoLLAMA3  \cite{videollama3}& 0.1371 & 0.1636 & 0.1016 & 0.2329 & 0.3048  & 0.1654 &  0.5082 & 0.5118 & 0.3862 & 0.2927 & 0.3267 & 0.2177 \\
  $\diamondsuit$ InternVL  \cite{internvl}& 0.1400 & 0.1589 & 0.0977 & 0.3470 & 0.3613 & 0.2427 & 0.5756 & 0.5916 & 0.4151 & 0.3542 & 0.3706 & 0.2518 \\
  $\diamondsuit$ mPLUG-OWl3 \cite{mplugowl3} & 0.2273 & 0.2129 & 0.1689 & 0.3237 & 0.3539 & 0.2381 & 0.2133 & 0.1779 & 0.1575 & 0.2548 & 0.2482 & 0.1882 \\
  \midrule
  $\bigstar$ HOST \cite{HOST} & 0.0846 & 0.0856 & 0.0568 & 0.0556 & 0.0599 & 0.0374 & 0.0440 & 0.0578 & 0.0293 & 0.0614 & 0.0678 & 0.0412 \\
  $\bigstar$ NIQE \cite{niqe} & 0.1031 & 0.0978 & 0.0698 & 0.0381 & 0.0455 & 0.0256 & 0.0621 & 0.1134 & 0.0418 & 0.0678 & 0.0856 & 0.0457 \\
  $\bigstar$ BRISQUE \cite{BRISQUE} & 0.1253 & 0.1005 & 0.0839 & 0.0352 & 0.0316 & 0.0235 & 0.0718 & 0.0661 & 0.0466 & 0.0774 & 0.0661 & 0.0513 \\
  $\bigstar$ BMPRI \cite{BMPRI} & 0.2731 & 0.2665 & 0.1853 & 0.0204 & 0.0277 & 0.0136 & 0.1568 & 0.1382 & 0.1030 & 0.1501 & 0.1441 & 0.1006 \\
  $\bigstar$ QAC \cite{qac} & 0.2847 & 0.2593 & 0.1924 & 0.0984 & 0.0973 & 0.0661 & 0.1385 & 0.1252 & 0.0901 & 0.1739 & 0.1606 & 0.1162 \\
  $\bigstar$ BPRI \cite{BPRI} & 0.3213 & 0.3210 & 0.2157 & 0.0223 & 0.0293 & 0.0150 & 0.2077 & 0.1884 & 0.1362 & 0.1838 & 0.1796 & 0.1223 \\
  \midrule
  $\spadesuit$ VSFA \cite{VSFA}& 0.7670 & 0.7522 & 0.5635 & 0.3775 & 0.3999 & 0.2603 & 0.7732 & 0.7755 & 0.5784 & 0.6392 & 0.6425 & 0.4674 \\
  $\spadesuit$ BVQA \cite{BVQA}& 0.7807 & 0.7810 & 0.5717 & 0.3215 & 0.3382 & 0.2984 & 0.7877 & 0.7863 & 0.5884 & 0.6300 & 0.6352 & 0.4862 \\
  $\spadesuit$ SimpleVQA \cite{simplevqa} & 0.7810 & 0.7846 & 0.5775 & 0.3670 & 0.3839 & 0.2511 & 0.7888 & 0.7902 & 0.5913 & 0.6456 & 0.6529 & 0.4733 \\
  $\spadesuit$ FAST-VQA \cite{fastvqa}  & 0.7925 & 0.7954 & 0.5861 & 0.3018 & 0.3229 & 0.2059 & 0.8001 & 0.8042 & 0.6021 & 0.6315 & 0.6408 & 0.4647 \\
  $\spadesuit$ DOVER \cite{dover}  & \textcolor{blue}{0.8096} & \textcolor{blue}{0.8155} & \textcolor{blue}{0.6203} & \textcolor{blue}{0.4279} & \textcolor{blue}{0.4684} & \textcolor{blue}{0.3017} & \textcolor{blue}{0.8010} & \textcolor{blue}{0.8056} & \textcolor{blue}{0.6149} & \textcolor{blue}{0.6795} & \textcolor{blue}{0.6965} & \textcolor{blue}{0.5123} \\
  \midrule
  \rowcolor{gray!20} \textbf{TDVE-Assessor (Ours)} & \textbf{\textcolor{red}{0.8688}} & \textbf{\textcolor{red}{0.8688}} & \textbf{\textcolor{red}{0.6919}} & \textbf{\textcolor{red}{0.8254}} & \textbf{\textcolor{red}{0.8330}} & \textbf{\textcolor{red}{0.6460}} & \textbf{\textcolor{red}{0.8354}} & \textbf{\textcolor{red}{0.8523}} & \textbf{\textcolor{red}{0.6564}} & \textbf{\textcolor{red}{0.8432}} & \textbf{\textcolor{red}{0.8514}} & \textbf{\textcolor{red}{0.6648}} \\
  \bottomrule
  \end{tabular}
}
\label{benchmark}
\end{table}
\vspace{-0.4cm}
\subsection{Experimental Setup}
\label{subsec:experimental_setup}
Performance is measured by the correlation between predicted scores and MOS using Spearman Rank Correlation Coefficient (SRCC), Pearson Linear Correlation Coefficient (PLCC), and Kendall Rank Correlation Coefficient (KRCC). We compared \textbf{TDVE-Assessor} against leading evaluation and metric models, categorized as follows:

Deep learning-based VQA models: DOVER \cite{dover}, FAST-VQA \cite{fastvqa}, SimpleVQA \cite{simplevqa}, VSFA \cite{VSFA}, BVQA \cite{BVQA}.
Traditional handcrafted models (Matlab-inferred): BMPRI, BPRI, BRISQUE, HOST, NIQE, QAC.
Text-visual consistency metrics: CLIPScore \cite{CLIPScore}, PickScore \cite{Kirstain2023PickaPicAO}, VQAScore \cite{vqascore}, BLIPScore, ImageReward, and AestheticScore.
Open-source Large Multimodal Models (LMMs): LLava-NEXT \cite{llavanext-video,liu2023llava,li2024llava,liu2024llavanext,liu2023improvedllava,li2024llavanext-strong,li2024llavanext-ablations}, VideoLLAMA3 \cite{videollama3,damonlpsg2023videollama,damonlpsg2024videollama2} , Internvideo2.5 \cite{internvideo}, Internvl \cite{internvl}, mPLUG-OWl3 \cite{mplugowl3}.

Traditional Non-Deep Learning Models are directly evaluated by averaging frame-level quality scores. Vision-language pre-trained frameworks and LMMs undergo zero-shot inference; metrics like CLIPscore \cite{CLIPScore} are derived from average cosine similarity. Fine-tunable VQA models (SimpleVQA \cite{simplevqa}, BVQA \cite{BVQA}, FAST-VQA \cite{fastvqa}, DOVER \cite{dover}) are domain-adapted on the respective test set distributions. We establish independent training pipelines per dimension for all deep learning methods, using a 4:1 train-validation split and averaging results over ten randomized trials.

\begin{figure}[t]
\vspace{-0.4cm}
\centering 
\setlength{\abovecaptionskip}{-0.2cm}
\setlength{\belowcaptionskip}{-0.4cm}
\includegraphics[width=\textwidth]{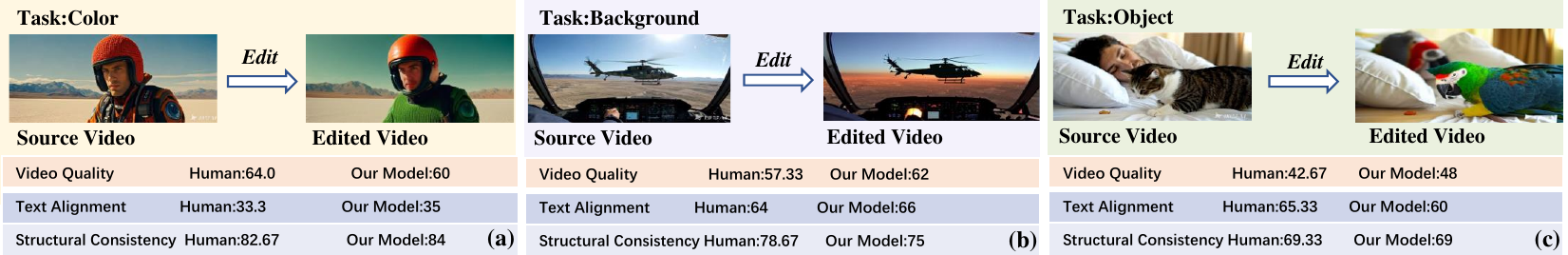} 
\caption{The example of the comparison between the actual score and the score predicted by our model. (a) Color Edit (b) Background Edit (c) Object Edit } 
\label{score_shili}
\end{figure}

\subsection{Performance Benchmarking}
\label{subsec:performance_benchmarking}
Table \ref{benchmark} presents the correlation results on \textbf{TDVE-DB}. Traditional NR-VQA \cite{zongshu} models performed poorly across all dimensions. Vision-language pre-training methods like CLIPscore \cite{CLIPScore} show moderate editing alignment performance but are surpassed by specialized, fine-tuned VQA models. While fine-tuned deep learning models (\textit{e.g.}, FAST-VQA \cite{fastvqa}, DOVER \cite{dover}) are more competitive, their overall performance remains unsatisfactory, particularly in editing alignment. This limitation often stems from their lack of direct text prompt input, hindering the extraction of text-video relational features. Notably, some LMMs demonstrate reasonable zero-shot performance, especially in structural consistency, benefiting from their flexible instruction-following capabilities. Crucially, our fine-tuned \textbf{TDVE-Assessor} achieves superior results across all three dimensions, underscoring its effectiveness.
\subsection{Cross-dataset evaluation}
\label{Cross-dataset evaluation}
To further demonstrate \textbf{TDVE-Assessor's} efficacy and generalization, we test it on VE-Bench DB, T2VQA-DB and AIGVQA-DB. It achieves state-of-the-art performance in both instances, as detailed in Table \ref{kuashujuji}. To qualitatively illustrate \textbf{TDVE-Assessor's} prediction capabilities, Figure \ref{score_shili} presents several examples comparing our model's prediction scores against actual human ratings across different editing tasks (\textit{e.g.}, color, background, and object edits). These examples showcase the model's ability to align with human perceptual judgments on all three dimensions.

\begin{table}[t]
\centering
\vspace{-6mm}
\caption{Benchmark on VE-Bench DB,T2VQA-DB and AIGVQA-DB}
\begin{adjustbox}{width=\textwidth}
\begin{tabular}{cccc|cccc|cccc}
    \toprule
    \multicolumn{4}{c}{\textbf{Benchmark on VE-Bench DB}} & \multicolumn{4}{c}{\textbf{Benchmark on T2VQA-DB}} & \multicolumn{4}{c}{\textbf{Benchmark on AIGVQA-DB}} \\
    \cmidrule(lr){1-4} \cmidrule(lr){5-8} \cmidrule(lr){9-12}
    Model & SRCC & PLCC & KRCC & Model & SRCC & PLCC & KRCC & Model & SRCC & PLCC & KRCC \\
    \midrule
    CLIP-F \cite{clip-f} & 0.2284 & 0.1860 & 0.1545 & UMTScore \cite{umtscore} & 0.0676 & 0.0721 & 0.0453 & VSFA \cite{VSFA} & 0.3365 & 0.3421 & 0.2268 \\
    $S_{edit}$  & 0.1686 & 0.1865 & 0.1135 & SimpleVQA \cite{simplevqa}& 0.6275 & 0.6338 & 0.4466 & BVQA \cite{BVQA} & 0.4594 & 0.4701 & 0.3268 \\
    PickScore \cite{Kirstain2023PickaPicAO}  & 0.2266 & 0.2446 & 0.1540 & BVQA \cite{BVQA}& 0.7390 & 0.7486 & 0.5487 & SimpleVQA \cite{simplevqa} & 0.8355 & 0.6438 & 0.8489 \\
    FAST-VQA \cite{fastvqa}& 0.6333 & 0.6326 & 0.4545 & FAST-VQA \cite{fastvqa}& 0.7173 & 0.7295 & 0.5303 & FAST-VQA \cite{fastvqa} & 0.8738 & 0.8644 & 0.6860 \\
    StableVQA \cite{stablevqa}& 0.6889 & 0.6783 & 0.4974 & DOVER \cite{dover}& 0.7609 & 0.7693 & 0.5704 & DOVER \cite{dover} & 0.8907 & 0.8895 & 0.7004 \\
    DOVER \cite{dover}& 0.6119 & 0.6295 & 0.4354 & T2VQA \cite{T2VQA} & 0.7965 & 0.8066 & 0.6058 & Q-Align \cite{qalign} & 0.8516 & 0.8383 & 0.6641 \\
    VE-Bench QA \cite{vebench}& 0.7415 & 0.7330 & 0.5414 & VE-Bench QA \cite{vebench} & 0.8179 & 0.8227 & 0.6370 & AIGV-Assessor \cite{AIGVASSESSOR} & 0.9162 & 0.9190 & 0.7576 \\
    \midrule
 \rowcolor{gray!20}   \textbf{TDVE-Assessor (Ours)} & \textbf{0.7527} & \textbf{0.7654} & \textbf{0.5645} & \textbf{TDVE-Assessor (Ours)} & \textbf{0.8222} & \textbf{0.8335} & \textbf{0.6399} & \textbf{TDVE-Assessor (Ours)} & \textbf{0.9397} & \textbf{0.9344} & \textbf{0.7835} \\
    \bottomrule
\end{tabular}
\end{adjustbox}
\label{kuashujuji}
\vspace{-1mm}
\end{table}

\subsection{Ablation Study}
\label{subsec:ablation_study}
Ablation results for score prediction are summarized in Table \ref{xiaorong}. Fine-tuning only the vision encoder (Exp.1) shows the weakest performance. Fine-tuning the LLM (Exp.2) significantly improves these results. Notably, incorporating the quality regression module with LLM fine-tuning (Exp.5 vs. Exp.2) markedly boosts performance. While fine-tuning both the vision encoder and LLM without regression (Exp.4) shows comparable results to fine-tuning only the LLM (Exp.2), configurations combining LLM fine-tuning and the regression module (Exp.5 and Exp.6) achieve the best overall scores. We select Exp.5 (LLM+Regression) as \textbf{TDVE-Assessor's} default due to its strong balance of performance and computational efficiency.

\begin{table}[t] 
  \centering       
  \caption{Comparison of Training Strategies} 

  \resizebox{\textwidth}{!}{
    \begin{tabular}{@{}lccc@{\quad}ccc@{\quad}ccc@{\quad}ccc@{}} 

      \toprule 

      & \multicolumn{3}{c}{Training Strategy} & \multicolumn{3}{c}{Video Quality} & \multicolumn{3}{c}{Editing Alignment} & \multicolumn{3}{c}{Structural Consistency} \\
      \cmidrule(lr){2-4} \cmidrule(lr){5-7} \cmidrule(lr){8-10} \cmidrule(lr){11-13}

      Exp & LoRA$_{r=8}$ (vision) & LoRA$_{r=8}$ (LLM) & Quality Regression & SRCC & PLCC & KRCC & SRCC & PLCC & KRCC & SRCC & PLCC & KRCC \\
      \midrule 

      Exp.1 & \checkmark &            &                    & 0.1445 & 0.1545 & 0.1031 & 0.3556 & 0.3718 & 0.3365 & 0.3874 & 0.3999 & 0.3381 \\
      Exp.2 &            & \checkmark  &                    & 0.7633 & 0.7604 & 0.6044 & 0.7968 & 0.8157 & 0.6424 & 0.7788 & 0.7901 & 0.6233 \\
      Exp.3 &            &            & \checkmark         & 0.6201 & 0.6276 & 0.4449 & 0.6177 & 0.6291 & 0.4485 & 0.6255  & 0.6331 & 0.4556 \\
      Exp.4 & \checkmark & \checkmark  &                    & 0.7641 & 0.7600 & 0.6056 & 0.7966 & 0.8155 & 0.6418 & 0.7785 & 0.7892 & 0.6222 \\
    \rowcolor{gray!20} \textbf{ Exp.5} &            & \checkmark  & \checkmark          & \textbf{0.8688} & \textbf{0.8688} &\textbf{ 0.6715} & \textbf{0.8254} & \textbf{0.8330} & \textbf{0.6460} &\textbf{ 0.8354} & \textbf{0.8523} & \textbf{0.6564} \\
      Exp.6 & \checkmark & \checkmark  & \checkmark         & 0.8679 & 0.8677 & 0.6909 & 0.8241 & 0.8321 & 0.6481 & 0.8321 & 0.8455 & 0.6547 \\

      \bottomrule 
    \end{tabular}
  } 
  \label{xiaorong}
  \vspace{-0.3cm}
\end{table}

\section{Conclusion}
\label{sub:conclusion}
This paper investigates text-driven video editing quality assessment. We first construct \textbf{TDVE-DB}, a large-scale, multi-dimensional benchmark dataset for video editing quality assessment, featuring a rich variety of editing categories and extensive human subjective ratings. Building upon this, we propose \textbf{TDVE-Assessor}, a novel LMM-based assessment model that innovatively combines video content and text instructions for multi-dimensional quality prediction. Experiments demonstrate that TDVE-Assessor achieves \textbf{State-of-the-Art (SOTA)} performance on both \textbf{TDVE-DB} and existing benchmark datasets, validating its effectiveness and strong generalization capabilities.

\textbf{Limitations and Societal Impact.} Our evaluation model, while demonstrating good generalization on TDVE-DB, is a baseline with scope for optimization, particularly as training focused on LLM text understanding and score regression. The current model rankings are based on data from selected professional annotators; thus, real-world application effectiveness remains an open question. We believe our benchmark and dataset will foster advancements in text-driven video editing, T2V generation, T2V evaluation, and V2T interpretation, encouraging further research.

{
    \small
    \bibliographystyle{ieeetr}
    \bibliography{neurips_2025}
}

\appendix
\part{Appendix} 

\section{ Ethical Discussions}
\subsection{Motivation and Potential Benefits}
We discuss how our work can benefit the community. Firstly, the primary motivation for our work is the rapid development of video editing technology, which now plays a significant role in various video clipping and creative video tasks. However, current video editing technologies \cite{tune-a-video,tokenflow,text2video-zero,ccedit,controlvideo,fatezero,flatten,fresco,pix2video,rave,slicedit,vid2vid} still face challenges, such as suboptimal implementation of multi-object interaction editing and difficulties with edits involving significant motion changes. Consequently, evaluating video editing technology has become a crucial task. There is a scarcity of existing quality assessment datasets for video editing; besides our work, TDVE-DB, only VE-Bench DB \cite{vebench} is available. Furthermore, there are currently few quality assessment methods \cite{vebench} specifically for text-driven video editing models. Due to the input-output characteristics of video editing, most existing models struggle to provide comprehensive and effective evaluations. Therefore, we propose TDVE-Assessor, a video quality assessment model based on Large Multimodal Models (LMMs) \cite{Qwen-VL}. By using TDVE-Assessor, we can accurately and effectively predict quality levels and scores for edited videos across three dimensions. Thanks to the inherent flexibility of LMMs \cite{Qwen-VL}, TDVE-Assessor exhibits strong generalization capabilities, achieving state-of-the-art (SOTA) performance on other Text-Driven Video Editing (TDVE) datasets and AI-Generated Video (AIGV) datasets. We believe that researchers aiming to explore video quality assessment in deeper, more complex dimensions can achieve excellent results by extending our work. Overall, we have established a novel and effective benchmark for the evaluation of text-driven video editing, pointing the way for future improvements in video editing.

\subsection{Ethical Discussions of Data Collection for TDVE-DB}
Regarding the construction of our dataset, TDVE-DB, the source videos are carefully selected. Real-world videos are obtained from publicly available datasets, and the AI-generated videos are used with permission for scientific research. All selected videos are free from copyright issues and do not contain sensitive content. For the generation of editing prompts, we employed an LLM for batch generation using a fixed prompt template. Following generation, these prompts underwent a manual review process to screen for and prevent any inappropriate or offensive content.

Prior to commencing the subjective experiments, all participants are fully informed about the intended use of their subjective scores. A screenshot of the user interface for the subjective experiment is shown in Figure 7. The approximate time commitment for each participant is 1.7 hours, and each participant received a compensation of \$11.50, a rate established in accordance with ethical standards.

\begin{figure}
    \centering
    \includegraphics[width=\linewidth]{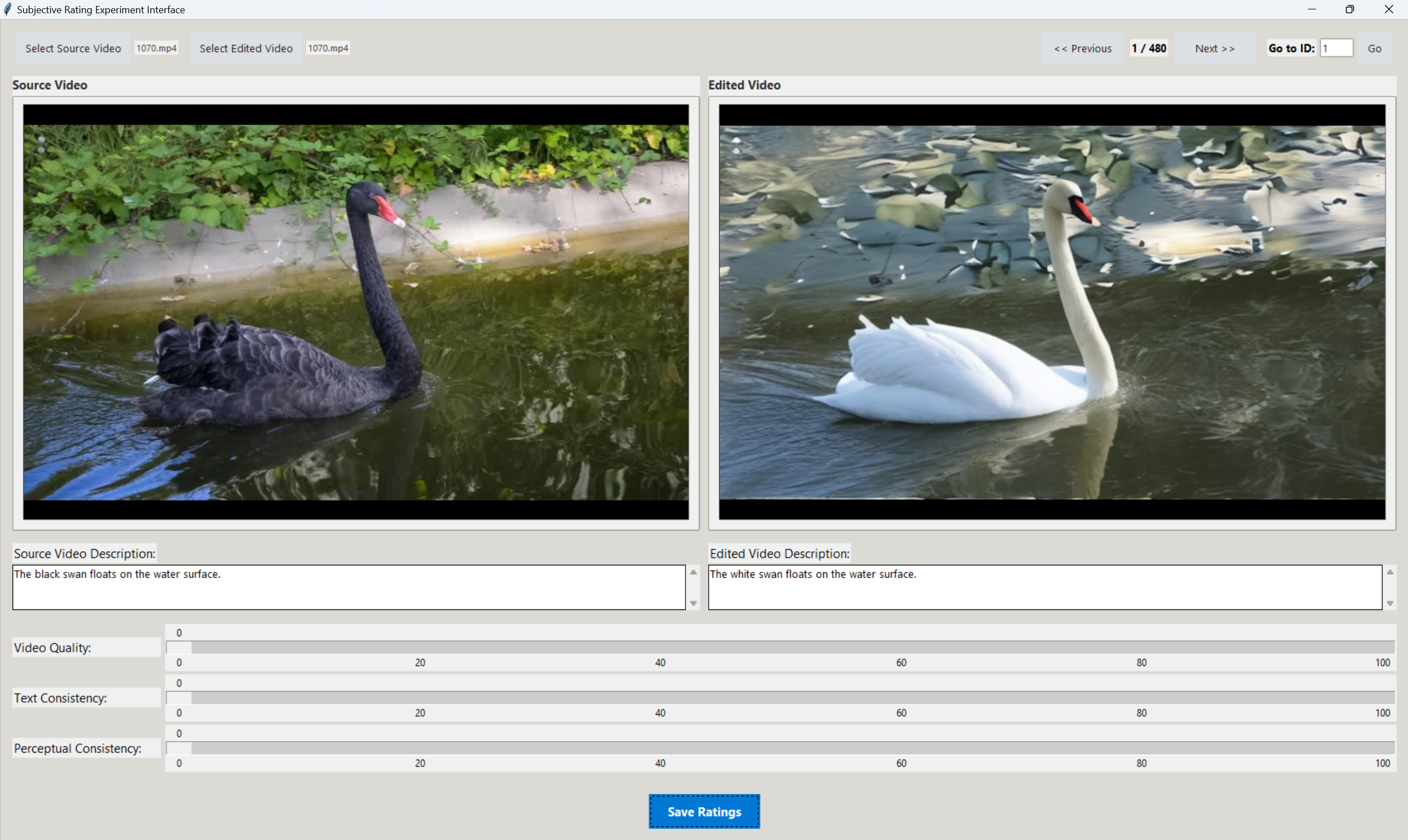}
    \caption{Subjective Experiment Rating UI.Participants evaluated the overall editing effect by considering the source video, the edited video, the source video description, and the edited video description. They provided scores across three dimensions: edited video quality, editing alignment, and structural consistency. }
    \label{fig:enter-label1}
\end{figure}
\section{More Details of TDVE-DB}
The project URLs for the text-driven video editing models used are listed in Table 6.

\subsection{Detailed Information of Text-Driven Video Editing Model}
\textbf{Tune-A-Video} \cite{tune-a-video}: Tune-A-Video is a one-shot tuning method for video editing that fine-tunes a pre-trained text-to-image (T2I) diffusion model to achieve text-driven video generation and editing. This method particularly focuses on maintaining temporal consistency in video content by extending 2D convolutions in T2I models to pseudo-3D convolutions and incorporating spatio-temporal information into the attention mechanism, requiring only a single video-text pair for tuning.

\textbf{Tokenflow} \cite{tokenflow}: Tokenflow proposes a text-driven video editing framework that utilizes pre-trained text-to-image diffusion models without requiring additional training or fine-tuning. Its core idea is to edit videos by enforcing consistency in the feature space of the diffusion model, specifically by explicitly propagating diffusion features based on inter-frame correspondences. This approach aims to preserve the original spatial layout and motion of the input video while adhering to text instructions.

\textbf{Text2Video-Zero} \cite{text2video-zero}: Text2Video-Zero is a zero-shot text-to-video (T2V) synthesis and editing model that can generate videos using existing text-to-image models like Stable Diffusion without any additional training or optimization. The model enriches the latent space of generated frames with motion dynamics to maintain global scene and background temporal consistency. It also utilizes a novel cross-frame attention mechanism (where each frame attends to the first frame) to preserve context, appearance, and structure, supporting text-to-video generation, pose/edge-guided generation, and instruction-guided video editing.

\textbf{CCEdit} \cite{ccedit} CCEdit (Creative and Controllable Video Editing) is a versatile generative video editing framework based on diffusion models. It employs a novel “trident network structure” that decouples structure and appearance control to achieve precise and creative editing. CCEdit utilizes the ControlNet architecture to maintain the structural integrity of the video during editing and incorporates an additional appearance branch to allow users fine-grained control over the edited keyframe, thereby balancing controllability and creativity.

\textbf{ControlVideo} \cite{controlvideo}: ControlVideo is a text-driven video editing method designed to generate videos that align with a given text while preserving the structure of the source video. Built upon pre-trained text-to-image diffusion models, it enhances the fidelity and temporal consistency of edited videos by incorporating additional conditional controls (such as edge maps or depth maps, often leveraging ControlNet concepts) and by optimizing keyframes and temporal attention mechanisms based on the source video-text pair. Some implementations support zero-shot editing and can be extended to handle long videos.

\textbf{FateZero} \cite{fatezero}: FateZero is a zero-shot text-driven video editing method that focuses on editing real-world videos without requiring per-prompt training or user-specific masks. Its core technique lies in attention fusion: it captures intermediate attention maps during DDIM (Denoising Diffusion Implicit Models) inversion to retain structural and motion information and directly fuses these maps into the editing process. Additionally, it uses cross-attention features from the source prompt to generate a blending mask for fusing self-attentions, which minimizes semantic leakage from the source video, thus achieving various edits like style transfer and local attribute modification while maintaining temporal consistency.

\textbf{FLATTEN} \cite{flatten}: FLATTEN (optical FLow-guided ATTENtion) is a training-free method designed to address content inconsistency in text-to-video editing. It introduces optical flow information into the attention module of the diffusion model's U-Net for the first time. By enforcing image patches that lie on the same optical flow path across different frames to attend to each other during the attention calculation, FLATTEN significantly improves the visual consistency of the edited videos. This method can be seamlessly integrated into various diffusion-based video editing workflows.

\textbf{FRESCO} \cite{fresco}: FRESCO (FRamE Spatial-temporal COrrespondence) is a zero-shot video translation and editing method that aims to extend image diffusion models to the video domain without additional training. It emphasizes that besides inter-frame temporal correspondence, intra-frame spatial correspondence is equally crucial for generating high-quality, coherent videos. FRESCO establishes more robust spatio-temporal constraints (by combining intra-frame and inter-frame correspondences) and explicitly updates features to achieve high spatio-temporal consistency with the input video, thereby enhancing the visual coherence of the edited videos.

\textbf{Pix2Video} \cite{pix2video}: Pix2Video is a method that utilizes pre-trained image diffusion models for video editing. It typically operates in a zero-shot or near zero-shot manner, applying image editing techniques to each frame of the video while focusing on addressing the challenge of temporal consistency across edited frames. By adapting the internal mechanisms or inputs of image diffusion models (such as specific versions of Stable Diffusion), Pix2Video aims to modify video content based on textual instructions while preserving natural transitions and coherent dynamics in unedited regions as much as possible.

\textbf{RAVE} \cite{rave}: RAVE (Randomized Noise Shuffling for Fast and Consistent Video Editing) is a zero-shot video editing method that leverages pre-trained text-to-image diffusion models. It edits videos based on text prompts while preserving the original video's motion and semantic structure, without requiring additional training. RAVE's core innovation is its novel “randomized noise shuffling strategy,” which efficiently utilizes spatio-temporal interactions between frames to produce temporally consistent videos. This approach is also designed for improved speed and memory efficiency, enabling it to handle longer video sequences.

\textbf{Slicedit} \cite{slicedit}: Slicedit introduces a zero-shot video editing method centered on the idea of using a pre-trained text-to-image (T2I) diffusion model to process “spatio-temporal slices” of a video. The method posits that these slices (pixel strips along the time axis) share statistical properties with natural images. Consequently, a powerful T2I model can be applied not only to individual frames (spatial slices) but also to these spatio-temporal slices. This allows for editing video content guided by text instructions while maintaining the original video's structure, motion, and the coherence of unedited regions.

\textbf{vid2vid-zero} \cite{vid2vid}: vid2vid-zero is a zero-shot video editing method designed to utilize off-the-shelf image diffusion models for editing video content without requiring model training or fine-tuning for specific tasks. The core of this approach lies in effectively transferring the strong generative priors of image models to the video domain. It employs specific techniques to adapt or guide these pre-trained image models to ensure that while edits are applied according to text instructions, the resulting video maintains temporal coherence and consistency across frames.

\begin{table}[t] 
    \centering 
    \caption{List of Text-Driven Video Editing Models and their URLs.}
    \label{tab:model_urls}
    \begin{tabular}{ll}
        \toprule 
        \textbf{Methods} & \textbf{URL} \\
        \midrule 
        Tune-A-Video \cite{tune-a-video} & \url{https://github.com/showlab/Tune-A-Video} \\
        Tokenflow \cite{tokenflow} & \url{https://github.com/omerbt/TokenFlow} \\
        Text2Video-Zero \cite{text2video-zero} & \url{https://github.com/Picsart-AI-Research/Text2Video-Zero} \\
        CCEdit \cite{ccedit} & \url{https://github.com/RuoyuFeng/CCEdit} \\
        ControlVideo \cite{controlvideo} & \url{https://github.com/YBYBZhang/ControlVideo} \\
        FateZero \cite{fatezero} & \url{https://github.com/ChenyangQiQi/FateZero} \\
        FLATTEN \cite{flatten} & \url{https://github.com/yrcong/FLATTEN} \\
        FRESCO \cite{fresco} & \url{https://github.com/williamyang1991/FRESCO} \\
        Pix2Video \cite{pix2video} & \url{https://github.com/duyguceylan/pix2video} \\
        RAVE \cite{rave} & \url{https://github.com/rehglab/RAVE} \\
        Slicedit \cite{slicedit} & \url{https://github.com/fallenshock/Slicedit} \\
        vid2vid-zero \cite{vid2vid} & \url{https://github.com/baaivision/vid2vid-zero} \\
        \bottomrule 
    \end{tabular}
\end{table}

\subsection{Source Video Characteristics}
\label{app:source_video_characteristics}

The TDVE-DB dataset is built upon 180 diverse source videos, which form the basis for the 3,857 edited video pairs. To ensure the dataset's breadth and representativeness, these videos are collected from two main categories:
\begin{itemize}
    \item \textbf{AI-Generated Videos}: This category comprises 68 videos. All are generated using Jimeng AI and have a uniform duration of 5 seconds.
    \item \textbf{Real-World Videos}: This category consists of 112 videos. These are carefully selected from the Kinetics-400 and DAVIS datasets, with original durations varying from 1 to 10 seconds. This selection aimed to cover a wide array of real-world scenarios and actions.
\end{itemize}

The \textbf{resolutions of the source videos are not uniform}. In the subsequent video editing process, the output resolution of some editing models follows that of the source video (indicated as ``Follow Source'' in Table 1 of the main paper), while other models may output videos at a fixed resolution. This diversity is intended to simulate the varied input conditions encountered in real-world video editing tasks. These sources are chosen to ensure a good balance and generalizability in terms of content and technical specifications.

\subsection{Detailed Description of Editing Prompts}
\label{app:editing_prompts_details}

The dataset includes 340 unique editing prompts, which are used to guide the 12 video editing models in generating the corresponding edited results. These prompts cover 8 distinct editing dimensions, such as color adjustment, motion modification, background replacement, and object editing.

The design of these prompts adhered to the following principles:
\begin{itemize}
    \item \textbf{Specificity and Feasibility}: To ensure the executability of the editing instructions and the accuracy of the evaluation, most editing prompts feature relatively minor semantic changes compared to the prompts describing the source video content. They typically focus on modifying a specific feature or attribute within the video.
    \item \textbf{Length and Complexity}: The vast majority of editing prompts are relatively short and concise. Only a small number of videos, approximately the first 10-20 in sequence, are associated with longer editing prompts, which naturally correspond to more complex editing instructions.
    \item \textbf{Objectivity and Constraints}: As illustrated in Figure 1(b) of the main paper, when generating prompts using Large Language Models (LLMs), we required the output editing instructions to be as objective as possible, avoiding subjective descriptions. We also aimed to limit the core editing content description to within 30 words. This constraint is primarily to ensure a clear focus for the edit. Most prompts meet this requirement. A few exceptions occur in the ``background replacement'' category, where the description of background content often necessitates more words, potentially causing some prompts to slightly exceed this limit.
\end{itemize}
\begin{figure}
    \centering
    \includegraphics[width=\linewidth]{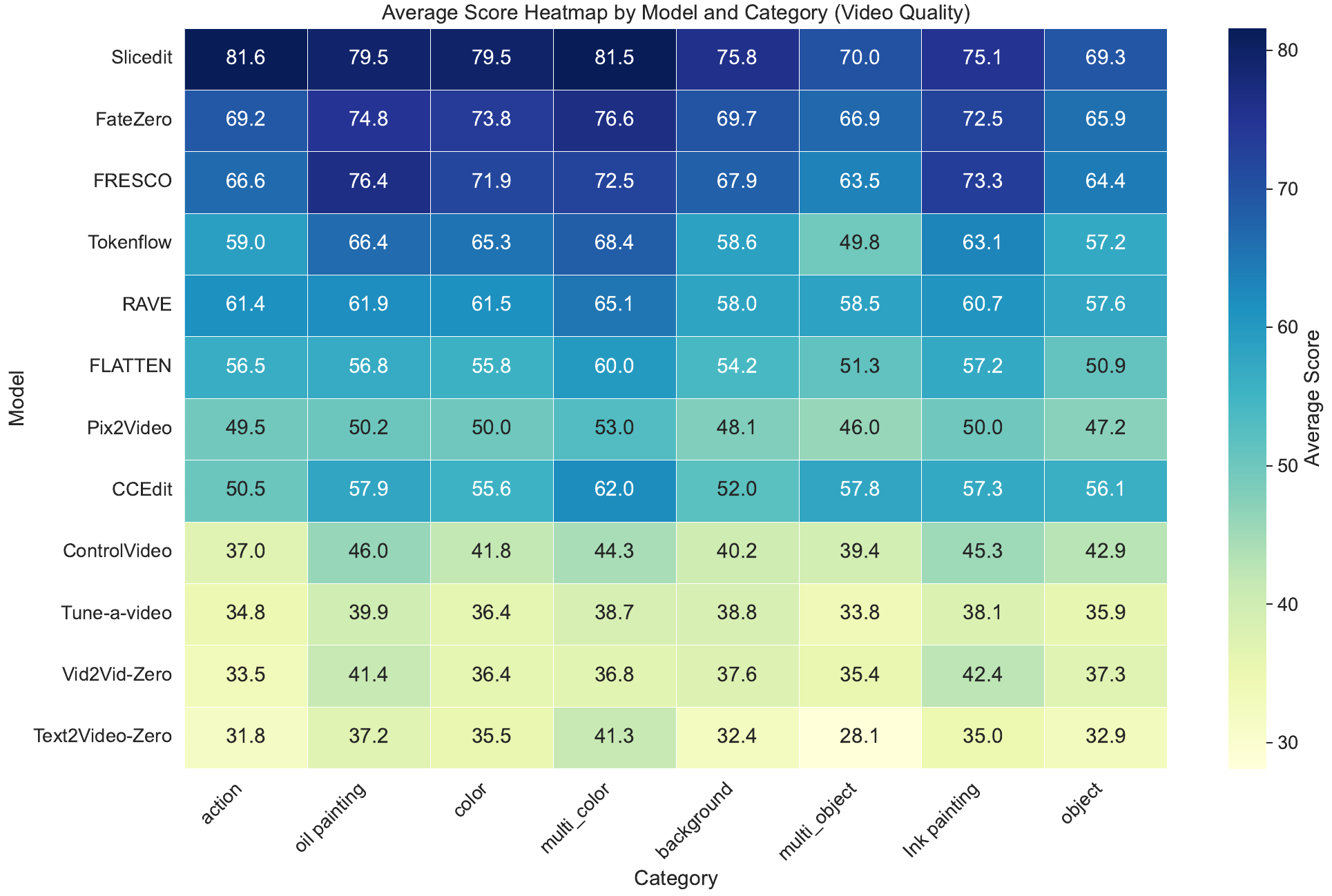}
    \caption{Video Quality}
    \label{fig:enter-label2}
\end{figure}
\begin{figure}
    \centering
    \includegraphics[width=\linewidth]{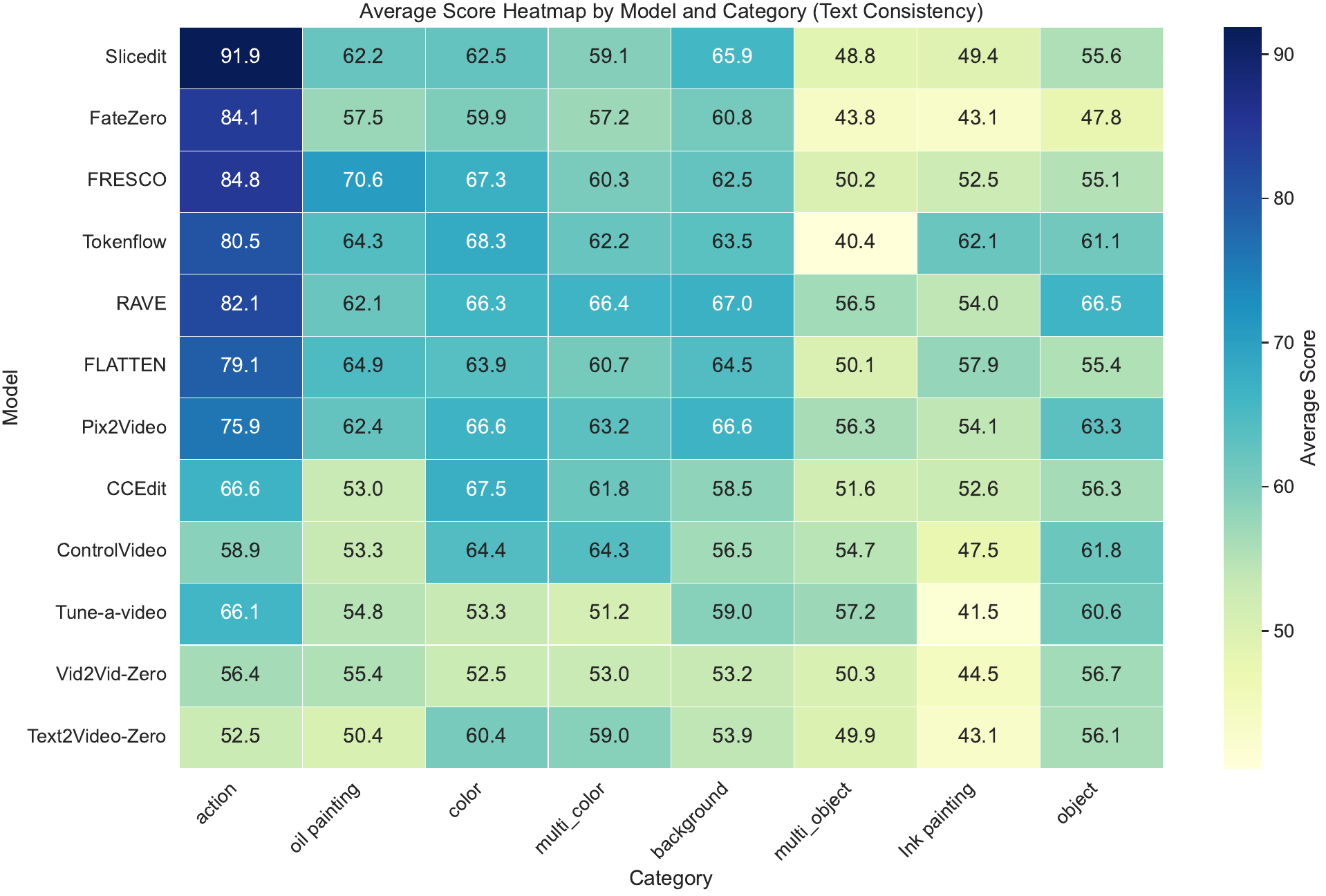}
    \caption{Editing Alignment}
    \label{fig:enter-label3}
\end{figure}
\begin{figure}
    \centering
    \includegraphics[width=\linewidth]{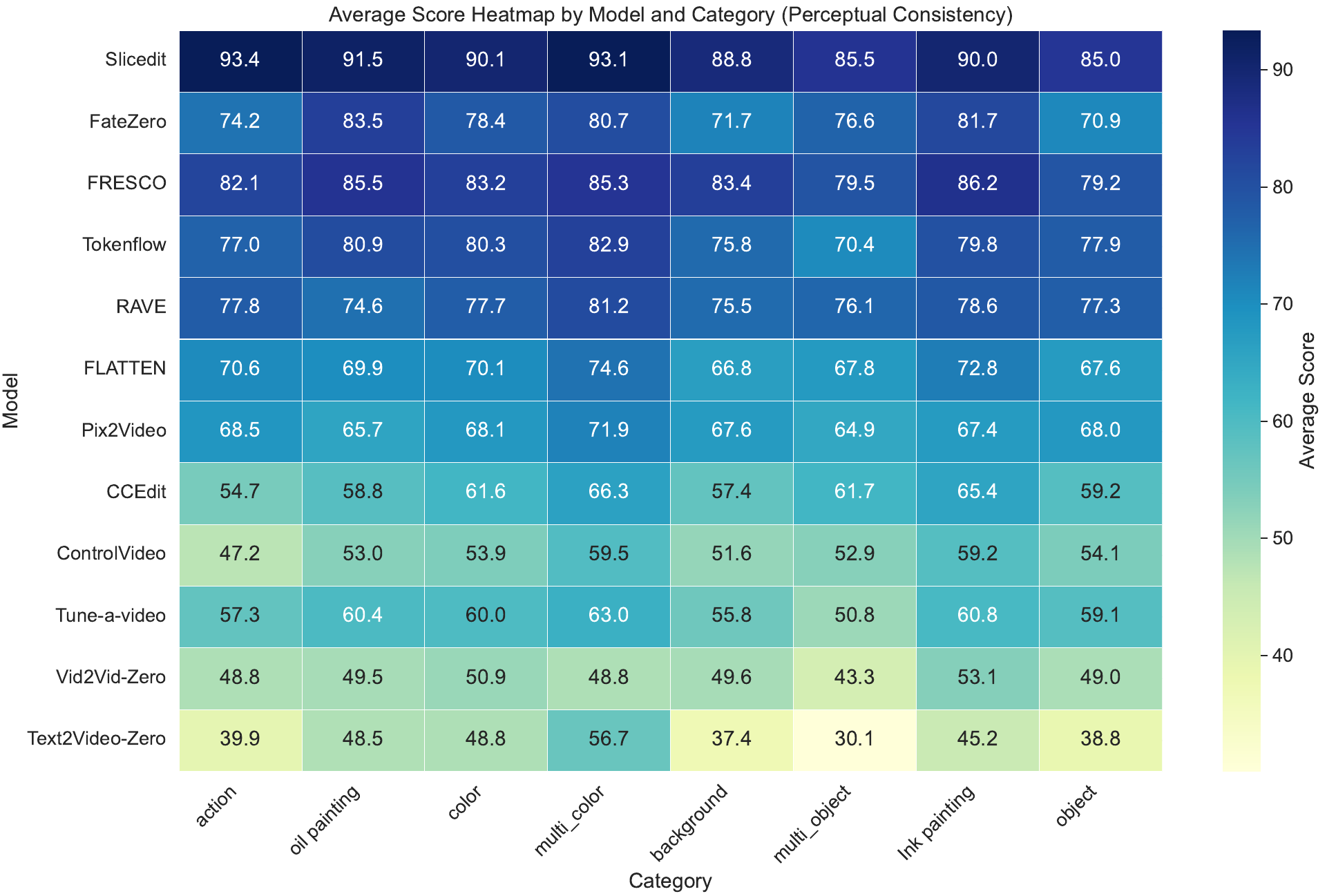}
    \caption{Structural Consistency}
    \label{fig:enter-label4}
\end{figure}
\begin{figure}
    \centering
    \includegraphics[width=\linewidth]{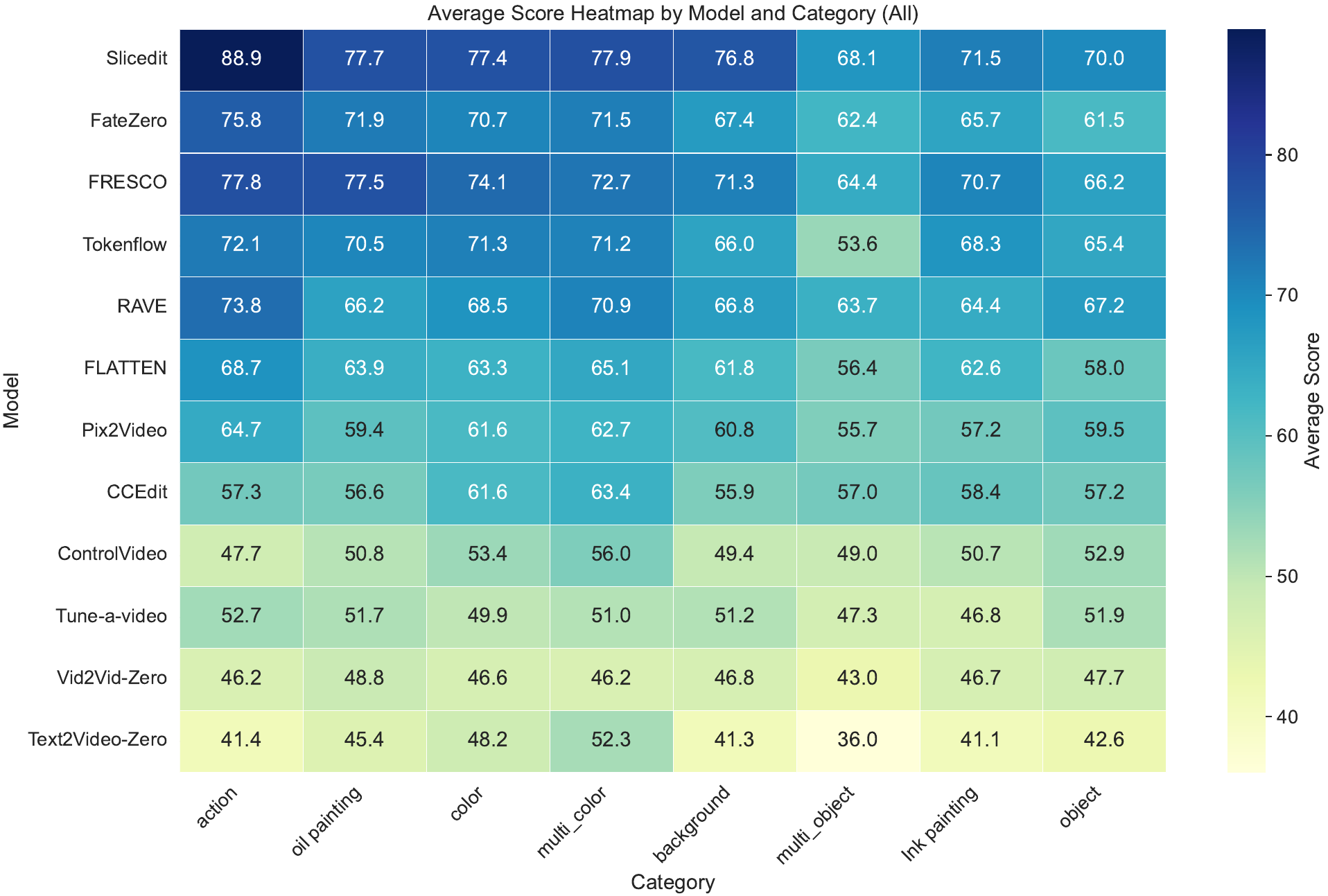}
    \caption{All Average}
    \label{fig:enter-label5}
\end{figure}
\subsection{Further Observations on Editing Categories and Score Distributions}
\label{app:categories_score_distribution}

The 8 editing categories covered in TDVE-DB exhibit certain differences in terms of editing difficulty and model performance:
\begin{itemize}
    \item \textbf{Editing Difficulty}: Based on subjective evaluations and model performance, ``color editing'' is generally considered one of the easier edit types to implement. Conversely, ``multi-object editing'' presents a greater challenge than ``single-object editing'', as the former requires managing more complex inter-object relationships and interactions while maintaining scene consistency.
    \item \textbf{Visualization of Score Distributions}: To more intuitively illustrate the performance of different editing models across various categories and evaluation dimensions, detailed score distributions are provided.
    \begin{itemize}
        \item Figure 2 presents bar charts showing the average scores for each model and editing category across the three dimensions.
        \item As a supplement, Appendix Figure 8 displays a heatmap of \textbf{Video Quality} scores for each model across different editing categories.
        \item Appendix Figure 9 displays a heatmap of \textbf{Editing Alignment} scores.
        \item Appendix Figure 10 displays a heatmap of \textbf{Structural Consistency} scores.
        \item Appendix Figure 11 displays a heatmap of the \textbf{Overall Average Score}, combining all three dimensions.
    \end{itemize}

\end{itemize}

\subsection{Qualitative Samples Displayed by Dimension and Quality Level}
For the convenience of the readers, we present the quality scores and quality levels of some video editing samples for reference, as shown in Figures 12, 13, and 14.
\begin{figure}
    \centering
    \includegraphics[width=0.88\linewidth]{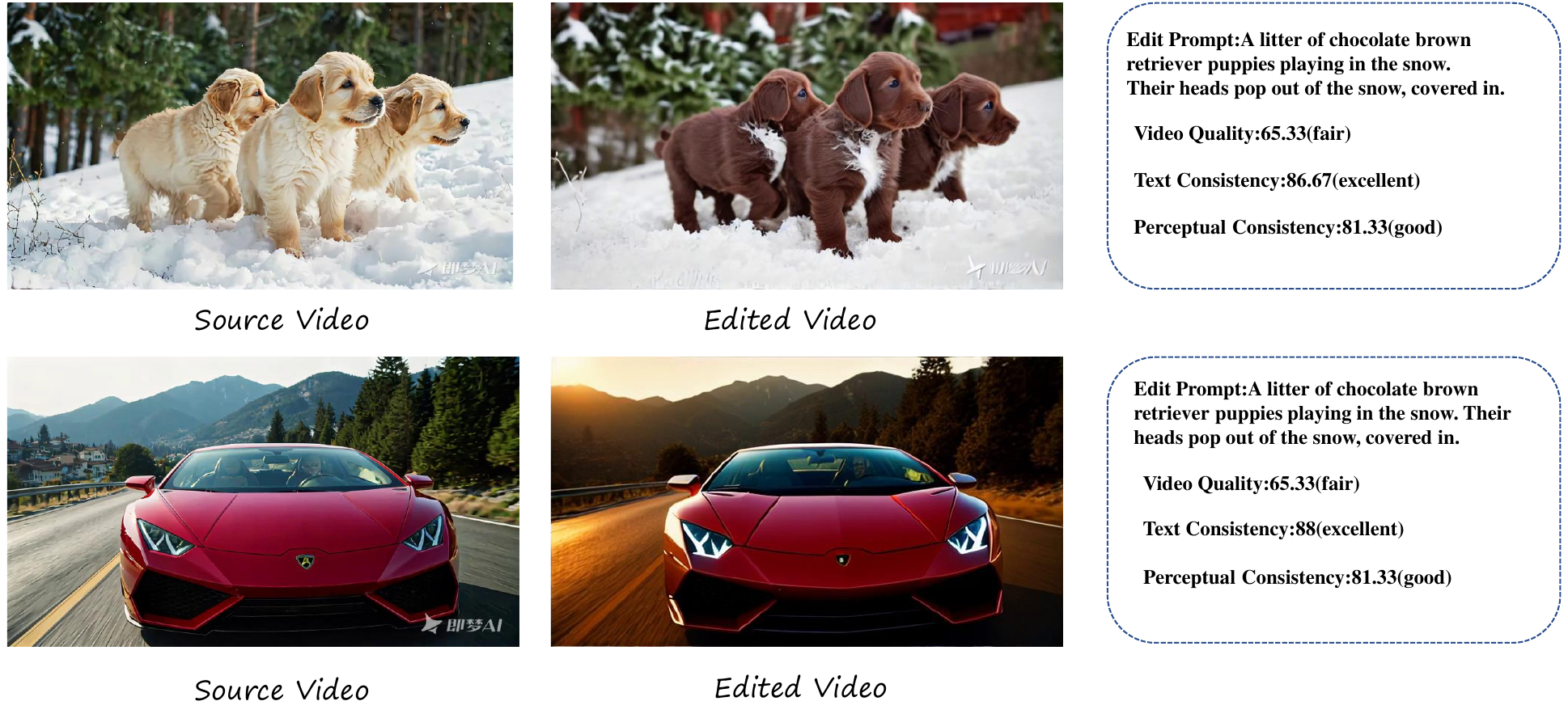}
    \caption{Sample 1}
    \label{fig:enter-label6}
\end{figure}

\begin{figure}
    \centering
    \includegraphics[width=0.88\linewidth]{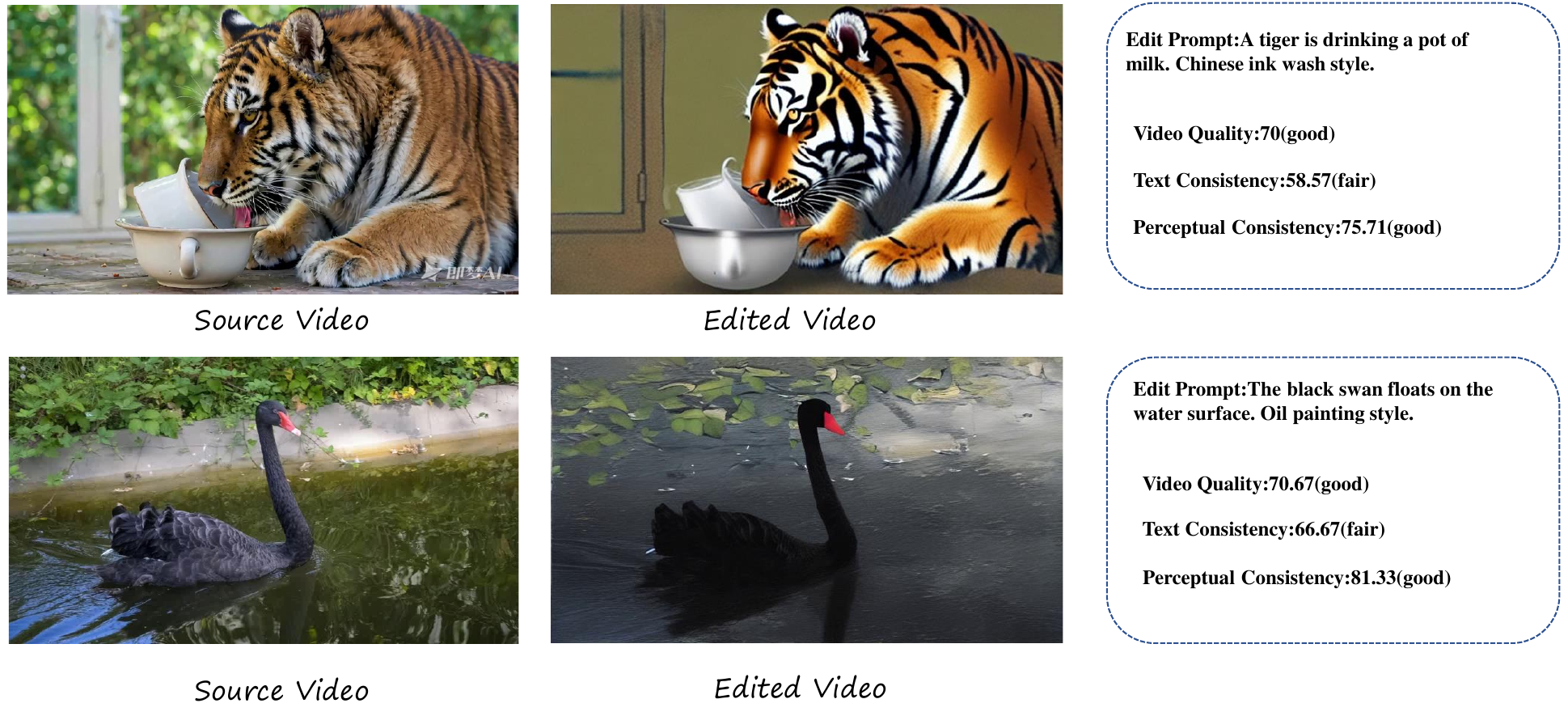}
    \caption{Sample 2}
    \label{fig:enter-label7}
\end{figure}

\begin{figure}
    \centering
    \includegraphics[width=0.88\linewidth]{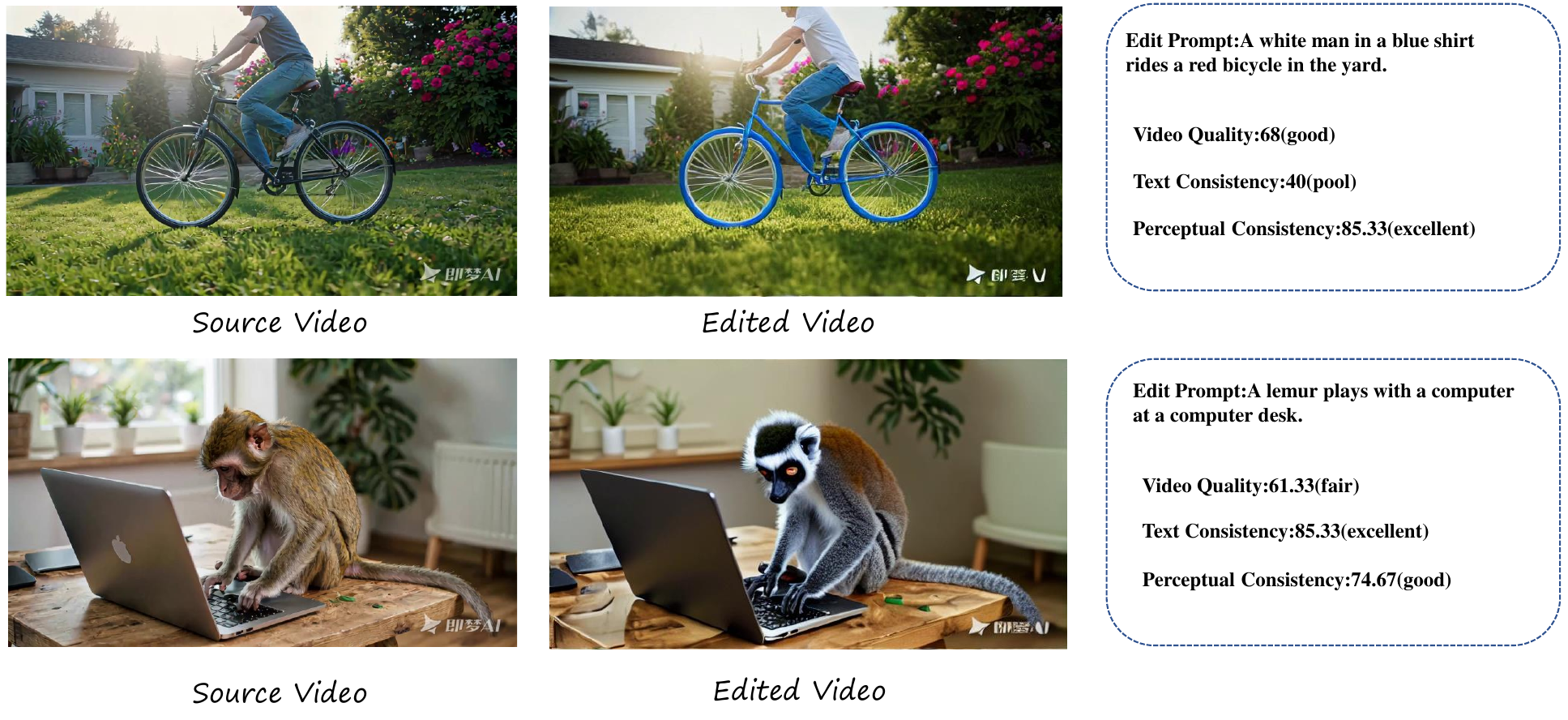}
    \caption{Sample 3}
    \label{fig:enter-label8}
\end{figure}

\subsection{Subjective Rating Reliability and Consistency Analysis}
\label{subsec:icc_analysis}

To ensure the reliability and consistency of the subjective scores in TDVE-DB, we conducted a rigorous inter-rater reliability (IRR) analysis on the 173,565 raw ratings. These ratings are collected from our 15 trained participants, as detailed in Section~\ref{subsec:Subjective Video Editing Quality Assessment} which describes our subjective evaluation methodology. We employed the Intraclass Correlation Coefficient (ICC) as the primary metric, specifically using a two-way random effects model for absolute agreement. This choice is well-suited for scenarios where raters are considered a representative sample from a larger population of potential raters, and it allows us to quantify the degree of consensus among them across the three core evaluation dimensions. The comprehensive results of this IRR analysis are presented and discussed below, with detailed statistics available in Table~\ref{tab:icc_results}.
\begin{table}[h!]
\centering
\caption{Inter-Rater Reliability (ICC) of Subjective Scores in TDVE-DB.}
\label{tab:icc_results}
\small 
\begin{tabularx}{\linewidth}{@{} l c c c c X X @{}} 
\toprule
\textbf{Evaluation Dimension} & \textbf{ICC2} & \textbf{95\% CI (ICC2)} & \textbf{ICC2k} & \textbf{95\% CI (ICC2k)} & \textbf{ICC2 Level*} & \textbf{MOS Reliability (ICC2k)} \\
\midrule
Video Quality           & 0.701         & (0.65, 0.75)  & 0.955          & (0.94, 0.97)   & Good          & Excellent                \\
Editing Alignment        & 0.753         & (0.71, 0.80)  & 0.920          & (0.89, 0.94)   & Excellent     & Excellent                \\
Structural Consistency  & 0.685         & (0.63, 0.74)  & 0.945          & (0.92, 0.96)   & Good          & Excellent                \\
\bottomrule
\end{tabularx}

\par 
\begin{minipage}{\linewidth} 
\footnotesize 
\vspace{0.5ex} 
\end{minipage}

\end{table}
The analysis indicates a commendable level of consistency and high reliability for the subjective scores within TDVE-DB:

\begin{enumerate}
    \item \textbf{Excellent Reliability of Mean Opinion Scores (MOS):} Across all three evaluation dimensions, the ICC2k values, which assess the reliability of the average scores from the 15 raters, are exceptionally high. Specifically, Video Quality (ICC2k = 0.955), Editing Alignment (ICC2k = 0.920), and Structural Consistency (ICC2k = 0.945) all significantly exceed the 0.90 threshold, signifying an “Excellent” level of reliability. This robustly confirms that the final MOS used as ground truth for benchmarking video editing models and for training our TDVE-Assessor are highly stable and dependable.

    \item \textbf{Consistency Among Individual Raters (ICC2):}
    \begin{itemize}
        \item For \textit{Editing Alignment}, the ICC2 value reached 0.753, signifying “Excellent” agreement among individual raters. This indicates that our participants achieved a strong consensus when evaluating the core semantic adherence of edited videos to the textual prompts.
        \item For \textit{Video Quality} (ICC2 = 0.701) and \textit{Structural Consistency} (ICC2 = 0.685), both ICC2 values fall into the “Good” range of agreement. These results reflect a solid level of consensus, even when assessing dimensions that inherently involve more nuanced individual perceptions of visual aesthetics and the subtle preservation of source video structure and content after editing.
    \end{itemize}
\end{enumerate}

\textbf{Discussion:}
The inter-rater reliability analysis strongly supports the quality and trustworthiness of the annotations within TDVE-DB. The excellent reliability of the Mean Opinion Scores (all ICC2k > 0.92) is paramount for its utility as a robust benchmark dataset. Furthermore, the good to excellent levels of agreement observed among individual raters (ICC2 values ranging from 0.685 to 0.753) underscore the clarity of our annotation guidelines and the rigorous training provided to our participants.

It is important to acknowledge the inherent complexity of evaluating text-driven video edits. Assessors are tasked with simultaneously considering multiple facets: the visual quality of the resultant video, its fidelity to the textual directive, and its consistency with the original source video's structural and content integrity. These multifaceted considerations, particularly those involving subjective aesthetic judgments and fine-grained semantic interpretations, can naturally introduce a degree of inter-rater variability. For instance, while “Editing Alignment” achieved an excellent ICC2, the “Video Quality” and “Structural Consistency” dimensions, which encompass broader visual and structural judgments, exhibited good, albeit slightly lower, ICC2 values. This reflects the inherent challenges in achieving perfect uniformity on these more holistic and sometimes subjective facets of video editing quality.

Despite these intrinsic challenges, the strong IRR results, especially the outstanding reliability of the aggregated MOS, confirm that TDVE-DB provides a dependable and valuable resource for the research community. The dataset effectively captures these complex human Structural nuances, offering a challenging and realistic platform for advancing both text-driven video editing technologies and the development of sophisticated VQA models, such as our proposed TDVE-Assessor.

\section{Details of Models Used in Benchmarking}

\subsection{Text-Visual Consistency Metrics}
\textbf{ImageReward} \cite{xu2023imagereward}: ImageReward is an assessment model based on human preference learning, specifically designed to evaluate the quality of text-to-image generation. It learns human preferences for image quality, aesthetics, and text-image alignment by training a reward model, thereby enabling scoring of generated images that is more consistent with human perception. Although primarily for images, its concept of learning human preferences can be relevant to the video domain.

\textbf{BLIPScore} \cite{li2022blip}: BLIPScore is an evaluation metric based on the BLIP (Bootstrapping Language-Image Pre-training) model, used to measure multimodal similarity between text and images. It utilizes a pre-trained vision-language model to understand image content and text descriptions, evaluating consistency by calculating their similarity in a shared embedding space, commonly used for tasks like image captioning evaluation.

\textbf{CLIPScore} \cite{CLIPScore}: CLIPScore is a widely used, reference-free evaluation metric for measuring the semantic similarity between images and text. It leverages the capabilities of the Contrastive Language-Image Pre-training (CLIP) model to encode images and text into the same multimodal embedding space, then calculates the cosine similarity between their embedding vectors as the consistency score. CLIPScore is extensively used in text-to-image generation and image captioning due to its effectiveness and convenience.

\textbf{PickScore} \cite{Kirstain2023PickaPicAO}: PickScore is also an evaluation metric based on human preference learning, aiming to better align the outputs of text-to-image models with human intuition. It is trained by collecting extensive user preference data on model-generated images, thereby learning a scoring function that can predict which images are more aligned with user expectations.

\textbf{VQAScore} \cite{vqascore}: VQAScore is a newer metric for evaluating the quality of text-to-visual content generation, which ingeniously utilizes the capabilities of image-to-text generation (\textit{e.g.}, image captioning or visual question answering). Its core idea is that if generated visual content is of high quality and consistent with the text prompt, a powerful vision-language model should be able to accurately answer questions related to the text prompt based on the visual content, or generate a description highly consistent with the prompt.

\textbf{AestheticScore} \cite{aesthe}: AestheticScore generally refers to models or metrics specifically designed to evaluate the aesthetic quality of images or videos. Such models learn features that distinguish high-aesthetic-quality visual content from low-quality content, which may be based on various visual elements like composition, color, and lighting. Some AestheticScore models are deep learning-based, trained on large datasets annotated with aesthetic ratings.

\subsection{Large Multimodal Models (LMMs)}
\textbf{LLaVA-NEXT} \cite{li2024llava}: LLaVA-NEXT is an improved version of the LLaVA (Large Language and Vision Assistant) series, aimed at enhancing the capabilities of large language models in visual understanding and multimodal dialogue. Through effective visual instruction tuning, LLaVA-NEXT can better follow user instructions, understand image or video content, and generate relevant text responses or perform visual question answering, showing potential for zero-shot video understanding.

\textbf{InternVideo2.5} \cite{internvideo}: The InternVideo series of models are foundational models focusing on general video understanding. InternVideo2.5, as a successor, further enhances performance on various video tasks such as video captioning, video question answering, and temporal action localization by combining generative and discriminative learning methods. The powerful video feature extraction and understanding capabilities of such models make them suitable for evaluating the semantic content and quality of videos.

\textbf{VideoLLAMA3} \cite{videollama3}: The VideoLLAMA series of models aims to extend the capabilities of large language models to the video domain, enabling deep understanding and interaction with video content. VideoLLAMA3, as its latest advancement, further enhances spatio-temporal modeling and audio understanding capabilities, allowing it to handle more complex video understanding tasks like long-video Q\&A and complex event reasoning.

\textbf{InternVL} \cite{internvl}: InternVL is a series of large-scale vision-language foundational models. Pre-trained on massive image-text data and aligned for various vision-language tasks, it excels in image understanding, visual question answering, and visual reasoning. These models are dedicated to building general-purpose vision-language representations that provide a strong foundation for downstream multimodal tasks.

\textbf{mPLUG-OWL3} \cite{mplugowl3}: The mPLUG-OWL series are pre-trained models for multimodal understanding and generation. mPLUG-OWL3 specifically focuses on improving the model's understanding of long image sequences (such as comics or long-form documents) and videos. Through innovative model architectures and pre-training tasks, it captures contextual information across images/frames, achieving good performance on long-sequence multimodal tasks.

\subsection{Traditional Handcrafted Image/Video Quality Assessment Models}
\textbf{HOST} \cite{HOST}: The HOST model, as described in Xu et al. (2016) (“Blind image quality assessment based on high order statistics aggregation”), assesses image quality by analyzing deviations in higher-order statistics (such as kurtosis and skewness) of image pixels or transform-domain coefficients, based on the premise that image distortions alter these natural statistical properties.

\textbf{NIQE} \cite{niqe} (Natural Image Quality Evaluator): NIQE is a no-reference image quality assessment model that operates by constructing a model from statistical features learned from high-quality natural images (Natural Scene Statistics - NSS). The quality of a test image is determined by measuring the deviation of its NSS features from this learned model of natural images.

\textbf{BRISQUE} \cite{BRISQUE} (Blind/Referenceless Image Spatial Quality Evaluator): BRISQUE is a no-reference image quality assessment method that extracts statistical features from locally normalized luminance coefficients in the spatial domain. These features are then used to train a regression model (Support Vector Regression - SVR) to predict subjective image quality scores.

\textbf{BMPRI} \cite{BMPRI}: The BMPRI method, as presented by Min et al. (2018) in “Blind image quality estimation via distortion aggravation,” estimates blind image quality through a process of distortion aggravation, analyzing how image characteristics change with further controlled degradation.

\textbf{QAC} \cite{qac}: The QAC approach, detailed in Xue et al. (2013) (“Learning without human scores for blind image quality assessment”), is a blind image quality assessment method that learns to predict image quality without relying on human subjective scores, potentially by formulating quality prediction as a classification task or learning from inherent statistical properties of images.

\textbf{BPRI} \cite{BPRI}: The BPRI method, from Min et al. (2018) (“Blind quality assessment based on pseudo-reference image”), is a blind image quality assessment technique that predicts image quality by first generating one or more pseudo-reference images and then comparing the image under assessment against these generated references.

\subsection{Deep Learning-based Video Quality Assessment (VQA) Models}

\textbf{VSFA} \cite{VSFA}: The VSFA model, introduced by Li et al. (2019) for “Quality assessment of in-the-wild videos,” is a deep learning-based VQA model that assesses video quality by analyzing spatio-temporal features, incorporating aspects like video saliency and fine-grained abnormality detection to predict quality scores for videos captured under unconstrained conditions.

\textbf{BVQA} \cite{BVQA}: In the context of your paper, BVQA  refers to the work by Tu et al. (2020), “Ugc-vqa: Benchmarking blind video quality assessment for user generated content.” This research provides a benchmark and dataset (UGC-VQA) for evaluating blind VQA models specifically on User Generated Content, addressing the unique challenges posed by such videos. The models and methodologies discussed or evaluated within this framework are pertinent here.

\textbf{SimpleVQA} \cite{simplevqa}: SimpleVQA , as proposed by Sun et al. (2022) in “A deep learning based no-reference quality assessment model for ugc videos,” is a specific no-reference video quality assessment model tailored for User Generated Content. It leverages pre-trained deep neural networks for spatio-temporal feature extraction, followed by a quality regression module, designed for simplicity and effectiveness.

\textbf{FAST-VQA} \cite{fastvqa} (Fast End-to-End Video Quality Assessment): FAST-VQA is an end-to-end video quality assessment model designed for efficiency. It employs techniques such as lightweight network architectures and fragment sampling to achieve rapid video quality assessment while maintaining accuracy.

\textbf{DOVER} \cite{dover}: DOVER, from Wu et al. (2023) (“Exploring video quality assessment on user generated contents from aesthetic and technical perspectives”), is a deep learning video quality assessment model that evaluates User Generated Content by considering both its aesthetic and technical qualities. It integrates diverse spatio-temporal features and may use methods like ordinal regression to align its predictions with human perceptual judgments.

\begin{table}[t] 
    \centering 
    \caption{Project URLs for Video Quality Assessment Method Benchmarked}
    \label{tab:bench_urls}
    \begin{tabular}{ll}
            \toprule 
        \textbf{Methods} & \textbf{URL} \\
        \midrule

ImageReward \cite{xu2023imagereward} & \url{https://github.com/THUDM/ImageReward} \\
BLIPScore \cite{li2022blip} & \url{https://github.com/salesforce/BLIP} \\
CLIPScore \cite{CLIPScore} & \url{https://github.com/jmhessel/clipscore} \\
PickScore \cite{Kirstain2023PickaPicAO} & \url{https://github.com/yuvalkirstain/PickScore} \\
VQAScore \cite{vqascore} & \url{https://github.com/linzhiqiu/t2v_metrics} \\
AestheticScore \cite{aesthe} & \url{https://github.com/sorekdj60/AestheticScore} \\
\midrule
LLaVA-NEXT \cite{llavanext-video} & \url{https://github.com/LLaVA-VL/LLaVA-NeXT} \\
InternVideo2.5 \cite{internvideo} & \url{https://github.com/OpenGVLab/InternVideo} \\
VideoLLAMA3 \cite{videollama3} & \url{https://github.com/DAMO-NLP-SG/VideoLLaMA3} \\
InternVL \cite{internvl} & \url{https://github.com/OpenGVLab/InternVL} \\
mPLUG-OWL3 \cite{mplugowl3} & \url{https://github.com/X-PLUG/mPLUG-Owl} \\ 

\midrule
VSFA \cite{VSFA} & \url{https://github.com/lidq92/VSFA} \\
BVQA \cite{BVQA} & \url{https://github.com/vztu/BVQA_Benchmark} \\
SimpleVQA \cite{simplevqa} & \url{https://github.com/sunwei925/SimpleVQA} \\
FAST-VQA \cite{fastvqa} & \url{https://github.com/timothyhtimothy/FAST-VQA-and-FasterVQA} \\
DOVER \cite{dover} & \url{https://github.com/VQAssessment/DOVER} \\
\bottomrule
    \end{tabular}
\end{table}

\section{Specific details of our model's training process}
\label{sec:appendix_training_details}

This section provides a detailed account of the training process for our TDVE-Assessor model. The training is conducted by separately addressing the three distinct assessment dimensions: video quality, text-video consistency, and Structural consistency. The LMM backbone used for TDVE-Assessor is Qwen2.5-VL-7B-Instruct.

\subsection{Hardware Configuration}
The training for all dimensions of the TDVE-Assessor is performed on a workstation equipped with two NVIDIA RTX A6000 GPUs, each with 48GB of VRAM. This configuration provided sufficient computational resources for handling the large multimodal model and the video data.

\subsection{Two-Stage Training Strategy}
The training of TDVE-Assessor is bifurcated into two distinct stages for each dimension. While the core methodology remains consistent across dimensions, the label format of the training data differed between stages, as detailed below. The 4:1 split for training and testing sets is kept identical for both stages.

\textbf{Stage 1: LLM Textual Semantic Prediction and Quality Classification Training}
\begin{itemize}
    \item \textbf{Objective:} To primarily train the Large Language Model (LLM) component for its text semantic prediction capabilities and to classify video quality into discrete levels, rather than predicting exact continuous scores.
    \item \textbf{Training Data Labels:} The ground truth labels provided to the model during this stage are textual quality classifications. For the video quality dimension, an example of the target output provided to the model is: \textit{``This quality of this video is poor.''}
    \item \textbf{Epochs:} 1 epoch.
\end{itemize}

\textbf{Stage 2: Quality Regression Module Training}
\begin{itemize}
    \item \textbf{Objective:} To train the quality regression module for predicting specific continuous quality scores, building upon the quality classification capability developed in Stage 1.
    \item \textbf{Training Data Labels:} The ground truth labels provided to the model during this stage included the specific numerical scores associated with the quality classifications. For the video quality dimension, an example of the target output is: \textit{``This quality of this video is poor (49.33).''}
    \item \textbf{Epochs:} 2 epochs.
\end{itemize}

\subsection{Dimension-Specific Prompting}
The primary distinction in training across the three dimensions (video quality, editing alignment, and structural consistency) lies in the prompts provided to the model:

\begin{itemize}
    \item \textbf{Video Quality Dimension:} The model is prompted to assess the standalone visual quality of the video. The prompt typically consisted of a direct question regarding the video's quality.
    \item \textbf{Editing Alignment Dimension:} The model is provided with the editing prompt (textual instruction) and the edited video. It is then prompted to assess the degree of alignment or consistency between the textual edit instruction and the resulting video content.
    \item \textbf{Structural Consistency Dimension:} For this dimension, two video segments (\textit{e.g.}, an original and an edited version, or two distinct temporal parts of a video intended to be consistent) are processed. Prior to model input, frames from these two video segments are sampled to an equal length and then concatenated (\textit{e.g.}, spatially side-by-side or temporally). The model is subsequently prompted to assess the spatial-structural consistency between the two presented video parts.
\end{itemize}

\subsection{Shared Training Hyperparameters}
The following hyperparameters are consistently used across all three dimensions during their respective training stages:

\begin{itemize}
    \item \textbf{Dataset Split:} The dataset is partitioned into training and testing sets using a 4:1 ratio.
    \item \textbf{Learning Rate:} $1 \times 10^{-4}$ (\textit{i.e.}, \texttt{1e-4}).
    \item \textbf{Batch Size:} 4.
    \item \textbf{LoRA Configuration:}
        \begin{itemize}
            \item LoRA Rank (\texttt{lora\_rank}): 8
            \item LoRA Alpha (\texttt{lora\_alpha}): 32
        \end{itemize}
    \item \textbf{Optimizer Settings:}
        \begin{itemize}
            \item Weight Decay (\texttt{weight\_decay}):
                \begin{itemize}
                    \item Stage 1: 0.01
                    \item Stage 2: 0.1
                \end{itemize}
            \item Warmup Ratio (\texttt{warmup\_ratio}): 0.05.
        \end{itemize}
\end{itemize}

This structured two-stage training approach, combined with dimension-specific prompting and carefully selected hyperparameters, allows TDVE-Assessor to effectively learn and predict nuanced aspects of text-driven video editing quality.


\end{document}